\documentclass[10pt,twocolumn,letterpaper]{article}

\usepackage{cvpr}
\usepackage{times}
\usepackage{epsfig}
\usepackage{graphicx}
\usepackage{amsmath}
\usepackage{subfigure}
\usepackage{amssymb}
\usepackage{booktabs}
\usepackage{algorithm}
\usepackage{algorithmic}
\usepackage[toc,page]{appendix}
\usepackage{diagbox}
\usepackage[normalem]{ulem}
\useunder{\uline}{\ul}{}

\usepackage{enumitem}

\usepackage[pagebackref=true,breaklinks=true,letterpaper=true,colorlinks,bookmarks=false]{hyperref}

\cvprfinalcopy % *** Uncomment this line for the final submission

\def\cvprPaperID{4856} % *** Enter the CVPR Paper ID here

\ifcvprfinal\fi
\begin{document}

%%%%%%%%% TITLE
\title{When NAS Meets Robustness: \\ In Search of Robust Architectures against Adversarial Attacks}
% \title{Towards Robust Network Architecture against Adversarial Attacks\\ with Neural Architecture Search}
% \title{Understanding the Effect of Network Architecture to Adversarial Robustness with Neural Architecture Search}

\author{
Minghao Guo\textsuperscript{1}\thanks{Equal contribution. Order determined by alphabetical order.} \quad
Yuzhe Yang\textsuperscript{2}\footnotemark[1] \quad\ 
Rui Xu\textsuperscript{1}\quad
Ziwei Liu\textsuperscript{1}\quad
Dahua Lin\textsuperscript{1}\vspace{.5em}\\
\textsuperscript{1}The Chinese University of Hong Kong \quad \textsuperscript{2}MIT CSAIL
% \vspace{-.5em}
}

\maketitle
%\thispagestyle{empty}

%%%%%%%%% ABSTRACT
\begin{abstract}
Recent advances in adversarial attacks uncover the intrinsic vulnerability of modern deep neural networks. Since then, extensive efforts have been devoted to enhancing the robustness of deep networks via specialized learning algorithms and loss functions. In this work, we take an \textbf{architectural perspective} and investigate the patterns of network architectures that are resilient to adversarial attacks. 
To obtain the large number of networks needed for this study, we adopt one-shot neural architecture search, training a large network for once and then finetuning the sub-networks sampled therefrom.
The sampled architectures together with the accuracies they achieve provide a rich basis for our study. 
Our ``robust architecture Odyssey'' reveals several valuable observations: 1) densely connected patterns result in improved robustness; 2) under computational budget, adding convolution operations to direct connection edge is effective; 3) flow of solution procedure (FSP) matrix is a good indicator of network robustness. Based on these observations, we discover a family of robust architectures (RobNets). 
On various datasets, including CIFAR, SVHN, Tiny-ImageNet, and ImageNet, RobNets exhibit superior robustness performance to other widely used architectures. Notably, RobNets substantially improve the robust accuracy ($\sim$5\% absolute gains) under both white-box and black-box attacks, even with fewer parameter numbers.
Code is available at {\small \url{https://github.com/gmh14/RobNets}}.
\end{abstract}

%%%%%%%%% BODY TEXT
\vspace{-.37cm}
\section{Introduction}
\vspace{-.03cm}
Deep neural networks are shown to be vulnerable to adversarial attacks, where the natural data is perturbed with human-imperceptible, carefully crafted noises~\cite{goodfellow2015explaining,kurakin2016adversarial,szegedy2013intriguing}.
To mitigate this pitfall, extensive efforts have been devoted to adversarial defense mechanisms, where the main focus has been on specialized adversarial learning algorithms \cite{goodfellow2015explaining,madry2017towards}, loss/regularization functions \cite{kannan2018adversarial,zhang2019theoretically}, as well as image preprocessing \cite{xie2017mitigating,song2018pixeldefend,xie2019feature,yang2019menet}. Yet, there is an orthogonal dimension that few studies have explored: \textbf{the intrinsic influence of network architecture on network resilience to adversarial perturbations}. 
Although the importance of architectures in adversarial robustness has emerged in the experiments of several previous work \cite{su2018robustness, xie2019intriguing, madry2017towards}, 
more comprehensive study on the role of network architectures in robustness remains needed.

% e.g. increasing the number of parameters can improve network robustness, 
\begin{figure}[tb]
\begin{center}
 \includegraphics[width=0.99\linewidth]{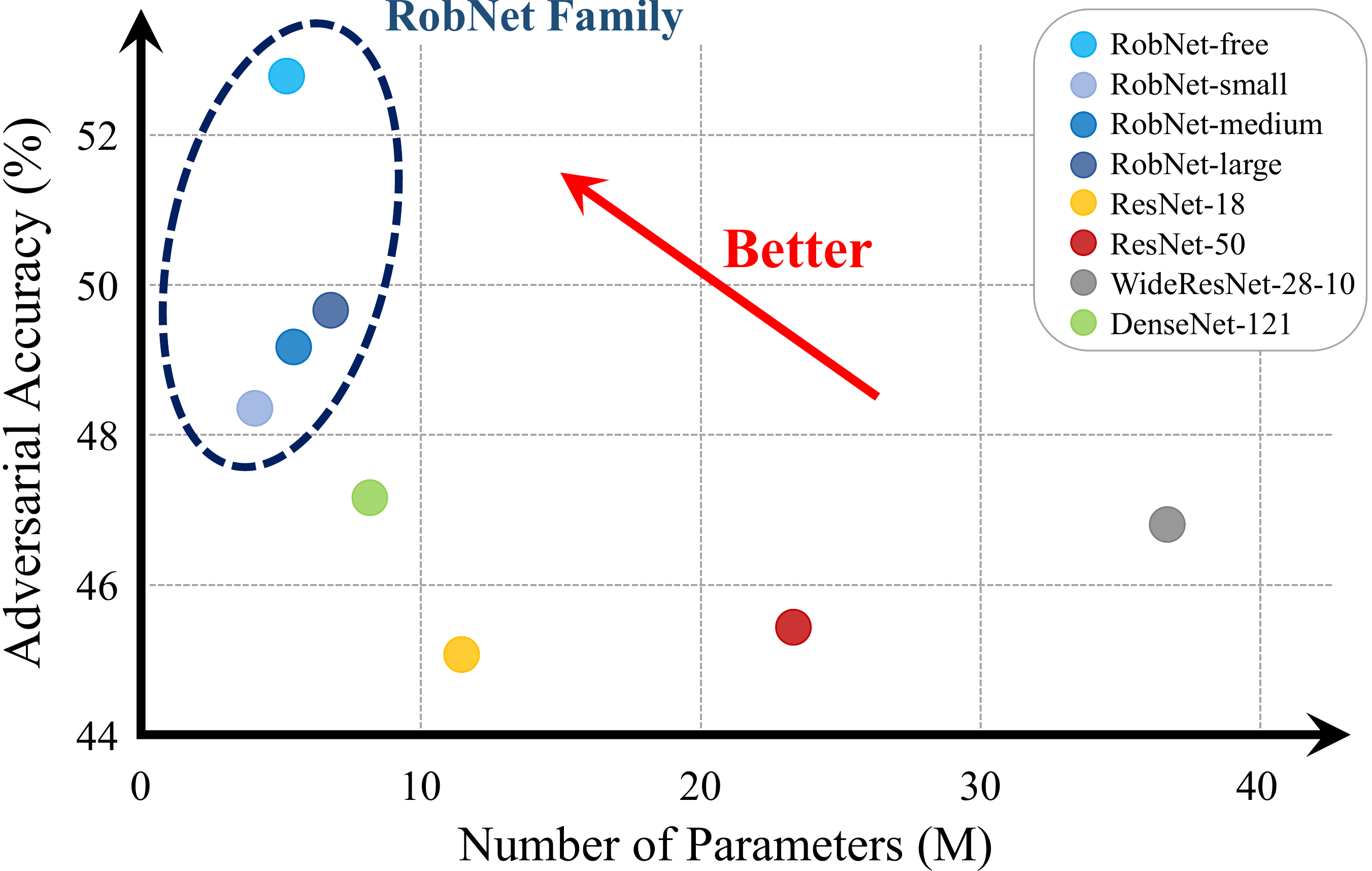}
\end{center}
\vspace{-0.2cm}
\caption{Adversarial robustness \emph{vs.} parameter numbers for widely used architectures and the proposed RobNet family on CIFAR-10. All models are adversarially trained using PGD with $7$ steps, and evaluated by PGD white-box attack with $100$ steps. RobNets exhibit superior robustness performance to other architectures, even with fewer parameter numbers.}
\label{fig-teaser}
\vspace{-0.2cm}
\end{figure}

In this work, we take the first step to systematically understand adversarial robustness of neural networks from an \emph{architectural perspective}. Specifically, we aim to answer the following questions: \label{sec-intro}
\begin{enumerate}[leftmargin=*]
\item \emph{What kind of network architecture patterns is crucial for adversarial robustness?}
\item \emph{Given a budget of model capacity, how to allocate the parameters of the architecture to efficiently improve the network robustness?}
\item \emph{What is the statistical indicator for robust network architectures?}
\end{enumerate}

It is nontrivial to answer the above questions, since we need to train a massive number of networks with different architectures and evaluate their robustness to gain insights, which, however, is exceedingly time-consuming, especially when adversarial training is used. 
Thanks to the method of one-shot neural architecture search (NAS), it becomes more accessible to evaluate robustness among a large number of architectures. Specifically, we first train a supernet for once, which subsumes a wide range of architectures as sub-networks, such as ResNet~\cite{He_2016_CVPR} and DenseNet~\cite{huang2017densely}. Then we sample architectures from the supernet and finetune the candidate architectures for a few epoches to obtain their robust accuracy under adversarial attacks. We further conduct extensive analysis on the obtained architectures and have gained a number of insights to the above questions: 

\textbf{1)} We present a statistical analysis on $1,000$ architectures from our cell-based search space, and discover a strong correlation between the density of the architecture and the adversarial accuracy. This indicates that densely connected pattern can significantly improve the network robustness.

\textbf{2)} We restrict the number of parameters under three different computational budgets, namely small, medium, and large. Our experimental results suggest that adding convolution operations to direct edges is more effective to improve model robustness, especially for small computational budgets.

\textbf{3)} We further release the cell-based constraint and produce studies on cell-free search space. For this setting, we find that the distance of flow of solution procedure matrix between clean data and adversarial data can be a good indicator of network robustness.

By adopting these observations, we search and design a family of robust architectures, called \emph{RobNets}. Extensive experiments on popular benchmarks, including CIFAR \cite{krizhevsky2009learning}, SVHN \cite{netzer2011reading}, Tiny-ImageNet \cite{tinyimagenet}, and ImageNet \cite{deng2009imagenet}, indicate that RobNets achieve a remarkable performance over widely used architectures.
% \textred{add detailed number against attacks? follow the flow of experiments}
Our studies advocate that future work on network robustness could concentrate more on the intrinsic effect of network architectures.

\section{Related Work}
\noindent
\textbf{Adversarial Attack and Defence.}
Deep neural networks (NNs) can be easily fooled by adversarial examples~\cite{goodfellow2015explaining,szegedy2013intriguing,liu2018dpatch}, where effective attacks are proposed such as FGSM \cite{goodfellow2015explaining}, BIM \cite{kurakin2016adversarial}, C\&W \cite{carlini2017towards}, DeepFool \cite{moosavi2016deepfool}, MI-FGSM \cite{dong2017boosting}, and PGD \cite{madry2017towards}.
Extensive efforts have been proposed to enhance the robustness, including preprocessing techniques \cite{buckman2018thermometer,song2018pixeldefend,yang2019menet}, feature denoising \cite{xie2019feature}, regularization \cite{zhang2019theoretically,kannan2018adversarial, lin2019defensive}, adding unlabeled data \cite{carmon2019unlabeled,stanforth2019labels}, model ensemble \cite{tramer2017ensemble,pang2019improving},
where adversarial training \cite{goodfellow2015explaining,madry2017towards} turns out to be the most effective and standard method for improving robustness.
A few empirical attempts on robustness of existing network architectures have been made~\cite{su2018robustness,xie2019intriguing}, but no convincing guidelines or conclusions have yet been achieved.

\vspace{.8em}
\noindent
\textbf{Neural Architecture Search.}
Neural architecture search (NAS) aims to automatically design network architectures to replace conventional handcrafted ones. Representative techniques include reinforcement learning \cite{zoph2016neural, zoph2017learning, zhong2018practical, guo2018irlas, baker2016designing}, evolution \cite{real2018regularized, suganuma2017genetic} and surrogate model \cite{liu2017progressive}, which have been widely adopted to the search process. However, these methods usually incur very high computational cost. Other efforts \cite{liu2018darts, perez2018efficient, bender2018understanding} utilize the weight sharing mechanism to reduce the costs of evaluating each searched candidate architecture. Towards a fast and scalable search algorithm, our investigation here is based on the one-shot NAS \cite{bender2018understanding, cai2019once}.

\section{Robust Neural Architecture Search}
\label{sec:robust-arch-search}
% In this section we first describe the core components of our NAS method for adversarial training. Then we produce comprehensive studies for the effect of cell-based architecture on network robustness under adversarial attack. \textred{(we need a short but impressive phase to summarize this sentence)} Lastly we release the cell-based constraint of architecture search space and conduct extensive analysis.
\subsection{Preliminary}
\label{sec-preliminary}

% \noindent \textbf{Motivation.}
% %
% In this work, we specifically examine the following questions: 

% \begin{enumerate}[leftmargin=*]
% \item \emph{What kinds of architecture pattern is crucial for the network robustness?}

% \item \emph{Given a budget of model capacity, how to allocate the parameters of the architecture for the maximum of the network robustness?}

% \item \emph{What is the long-range statistical indicator for network robustness across layers?}
% \end{enumerate}
% It is non-trivial to answer the above question, since we need to train a massive variety of architectures and evaluate their robustness to gain insights for the characteristic of robust architectures. Training such an amount of network architectures individually is exceedingly time-consuming and computationally costing. The case goes worse when we apply adversarial training techniques to obtain the maximal robustness of each network. Thanks to the method of one-shot NAS, it becomes more accessible to evaluate robustness of a large number of architectures. 
In this section, we briefly introduce the concept of one-shot NAS and adversarial training for the ease of better understanding of our further analysis.

\begin{figure*}[h]
\centering
\includegraphics[width=0.95\linewidth]{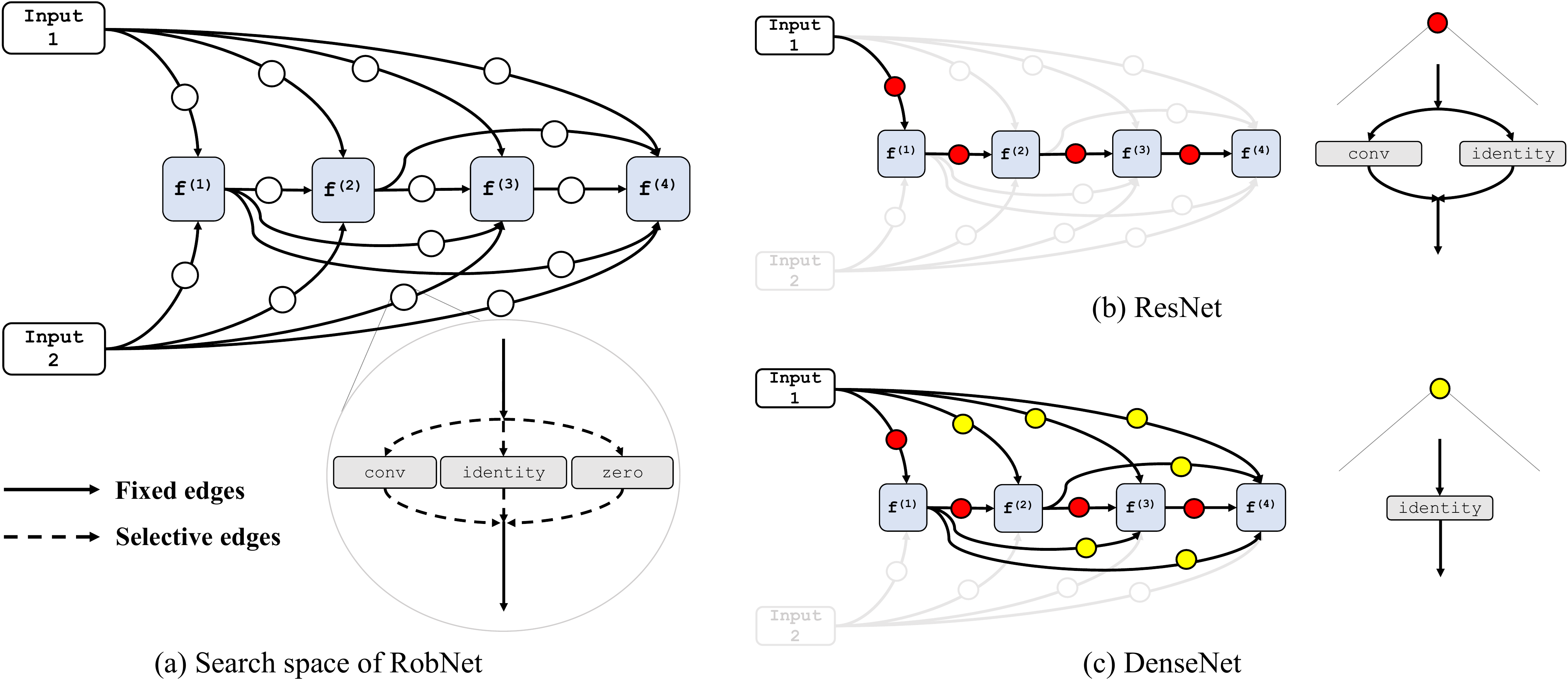}
% \vspace{-0.05cm}
\caption{Overview of the search space in robust architecture search: (a) The search space of RobNet. We only consider three candidate operations: $3\times3$ separable convolution, identity, and zero. We do not restrict the number of edges between two intermediate nodes to be one, which means that there could be multiple operations between two nodes. Such design benefits us to explore a larger space with more variants of network topology, including many typical human-designed architectures such as (b) ResNet and (c) DenseNet.}
\label{fig-res-densenet}
\vspace{-0.1cm}
\end{figure*}

\vspace{.8ex}
\noindent \textbf{One-Shot NAS.}
The primary goal of NAS \cite{zoph2017learning, liu2017progressive, liu2018darts, bender2018understanding} is to search for computation cells and use them as the basic building unit to construct a whole network. 
The architecture of each cell is a combination of operations chosen from a pre-defined operation space. 
In one-shot NAS, we construct a \emph{supernet} to contain every possible architecture in the search space. We only need to train the supernet for once and then at evaluation time, we can obtain various architectures by selectively zero out the operations in the supernet. The weights of these architectures are directly inherited from the supernet, suggesting that weights are \emph{shared} across models. The evaluated performance of these one-shot trained networks can be used to rank architectures in the search space, since there is a near-monotonic correlation between one-shot trained and stand-alone trained network accuracies. We refer readers to \cite{bender2018understanding} for more details of this order preservation property.

In one-shot NAS, the one-shot trained networks are typically only used to rank architectures and the best-performing architecture is \emph{retrained} from scratch after the search. In our work, however, we do not aim to get one \emph{single} architecture with the highest robust accuracy, but to study the effect of different architectures in network robustness. Thus, we do not involve retraining stage in our method but fully utilize the property of accuracy order preservation in one-shot NAS.

\vspace{.8ex}
\noindent \textbf{Robustness to Adversarial Examples.}
%
% which is the key of the improvement in efficiency comparing to training each network separately. 
Network robustness refers to how network is resistant to adversarial inputs. The problem of defending against bounded adversarial perturbations can be formulated as follows:
\begin{eqnarray}
\min_{\theta}\ \mathbb{E}_{(x,y)\sim\mathcal{D}}\left[\max_{x'\in S}\ \mathcal{L}\left(y, M(x';\theta)\right)\right], \label{eq-adv-training}
\end{eqnarray}
where $S = \{x': ||x-x'||_{p}<\epsilon\}$ defines the set of allowed perturbed inputs within $l_p$ distance, $M$ denotes the model and $\mathcal{D}$ denotes the data distribution.
One promising way to improve network robustness is adversarial training. \cite{madry2017towards} proposed to use Projected Gradient Descent~(PGD) to generate adversarial examples and augment data during training, which shows significant improvements of network robustness. 
% PGD adversarial examples are generated using the following updates:
% \begin{eqnarray}
% x^{t+1} = \Pi_{S}(x^t + \eta\cdot {\rm{sign}}(\nabla_{x}\mathcal{L}_{\rm{CE}}(\theta,x^t,y)), 
% \end{eqnarray}
% where $\eta$ denotes the step size and $t$ is the index of the step number.
In our study, we focus on adversarial attacks bounded by $l_\infty$ norm and use PGD adversarial training to obtain robust networks of different architectures.

\subsection{Robust Architecture Search Framework}
We now describe the core components of our robust architecture search framework. Our work is based on conventional one-shot architecture search methods \cite{bender2018understanding, liu2018darts}, with certain modifications to facilitate adversarial training and further analysis. We introduce them in detail accordingly.

\vspace{.8ex}
\noindent \textbf{Search Space.} 
% \textred{This paragraph is not very smooth}
Following \cite{zoph2017learning,liu2017progressive,liu2018darts,bender2018understanding}, we search for computation cell as the basic building unit to construct the whole network architecture. Each cell is represented as a directed acyclic graph $G=(V,E)$ consisting of $N$ nodes. Each node $V^{(i)}$ corresponds to a intermediate feature map $f^{(i)}$. Each edge $E^{(i,j)}$ represents a transformation $o^{(i,j)}(\cdot)$ chosen
from a pre-defined operation pool $\mathcal{O}=\{o_k(\cdot), k=1,\dots,n\}$ containing $n$ candidate operations (see Fig.~\ref{fig-res-densenet}(a)). The intermediate node is computed based on all of its predecessors: $f^{(j)}=\sum_{i<j}o^{(i,j)}(f^{(i)})$. The overall inputs of the cell are the outputs of previous two cells and the output of the cell is obtained by applying concatenation to all the intermediate nodes. For the ease of notation, we introduce architecture parameter $\alpha=\{\alpha_k^{(i,j)}| \alpha_k^{(i,j)}\in\{0,1\},i,j=1,\dots,N, \ k=1,\dots,n\}$ to represent candidate architectures in the search space. Each architecture corresponds to a specific architecture parameter $\alpha$. For edge $E^{(i,j)}$ of an architecture, the transformation can then be represented as $o^{(i,j)}(\cdot)=\sum_{k=1}^{n}\alpha^{(i,j)}_k o_k(\cdot)$. 
We refer \emph{direct edges} to those $E^{(i,j)} \in \{E^{(i,j)}|\ j=i+1\}$ and refer \emph{skip edges} to those $E^{(i,j)} \in \{E^{(i,j)}|\ j>i+1\}$. 

The main differences in search space between our work and conventional NAS lie in two aspects: 1) We shrink the total number of candidate operations in $\mathcal{O}$, remaining only: $3\times3$ separable convolution, identity, and zero. This helps to lift the burden of adversarial training, while remaining powerful candidate architectures in the search space \cite{xie2019exploring}. 2) We do not restrict the maximal number of operations between two intermediate nodes to be one (i.e., $o^{(i,j)}(\cdot)$ could contain at most $n$ operations). 
As shown in Fig.~\ref{fig-res-densenet}, such design encourages us to explore a larger space with more variants of network architectures, where some classical human-designed architectures can emerge such as ResNet and DenseNet. 

\begin{figure}[tb]
\centering
\subfigure[]{\includegraphics[width=.49\linewidth]{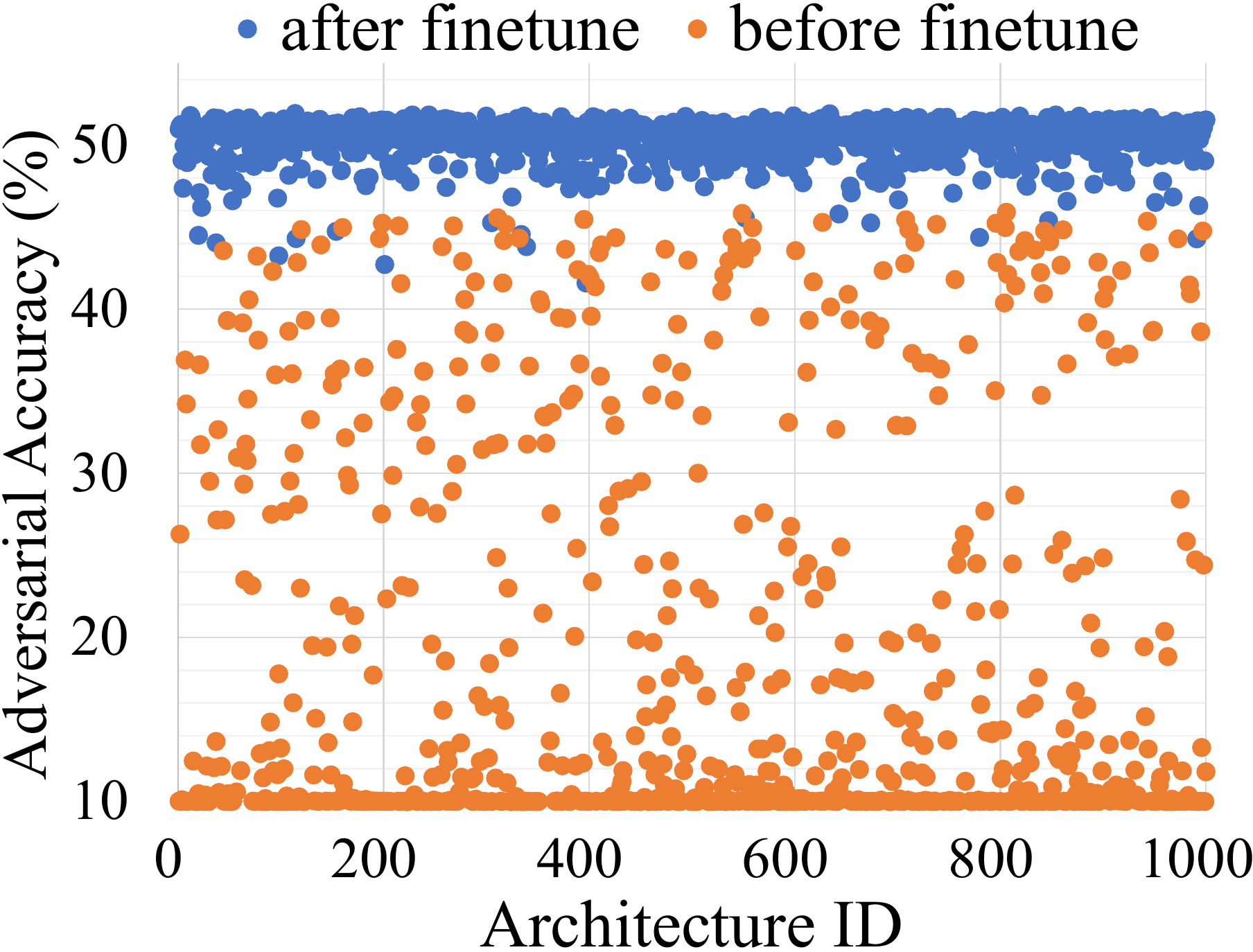}\label{fig-acc-improve}}
\subfigure[]{\includegraphics[width=.49\linewidth]{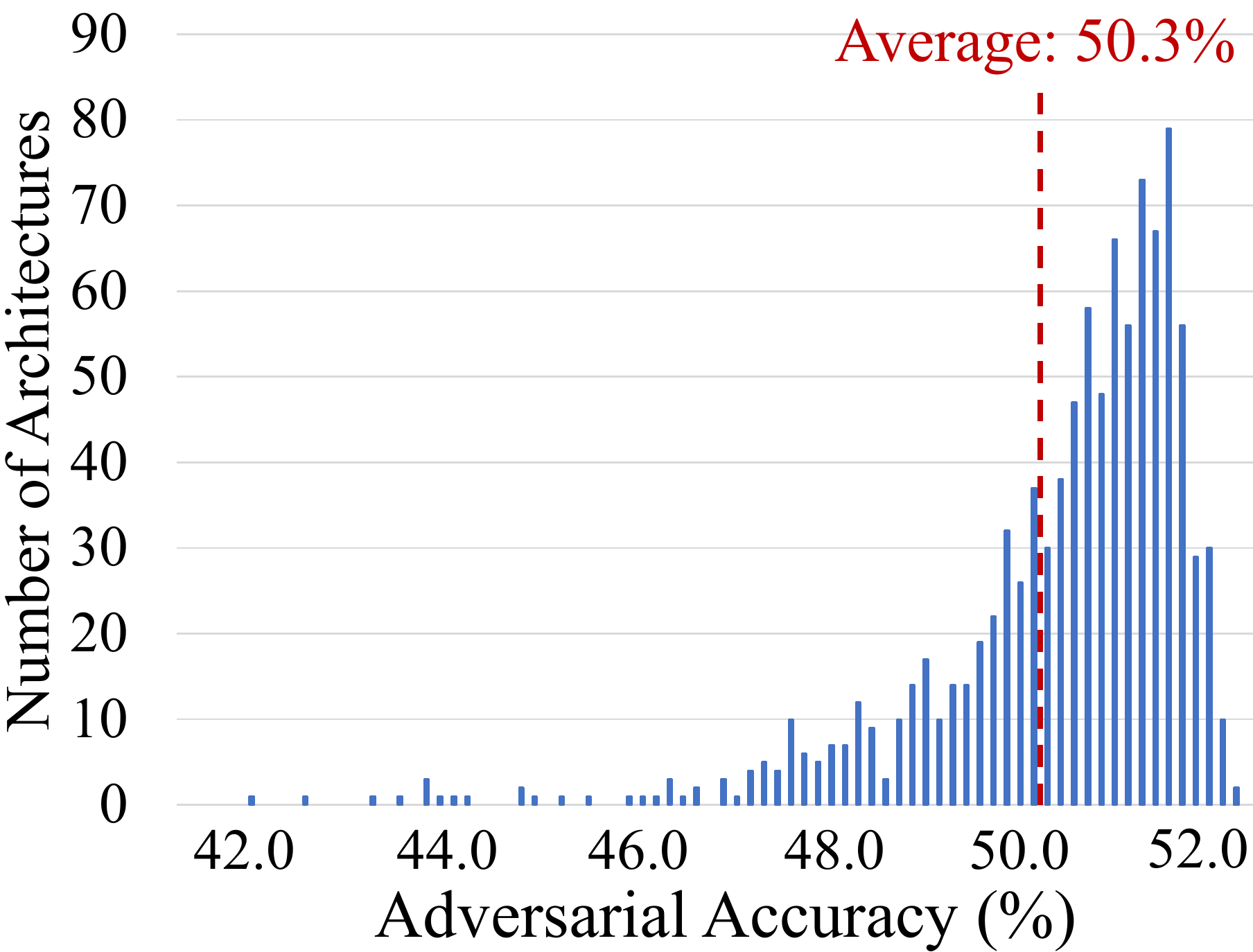}\label{fig-hist}}
\caption{Statistical results over 1,000 sampled architectures: (a) The robust accuracy improvement before and after finetuning. (b) Histogram of adversarial accuracy for all sampled architectures.}
\vspace{-.2cm}
\end{figure}

\vspace{.8ex}
\noindent \textbf{Robust Search Algorithm.}
\label{search-algorithm}
% more clear and detail description
We develop our robust search algorithm based on the one-shot NAS approach \cite{bender2018understanding}. Specifically, we set all the elements in architecture parameter $\alpha$ as 1 to obtain a supernet containing all possible architectures. During the training phase of the supernet, for each batch of training data, we randomly sample a candidate architecture from the supernet (by arbitrary setting some of the elements in $\alpha$ to 0). This path dropout technique is incorporated to decouple the co-adaptation of candidate architectures \cite{bender2018understanding}.
We then employ the min-max formulation in Eq.~(\ref{eq-adv-training}) to generate adversarial examples with respect to the sampled sub-network, and perform adversarial training to minimize the adversarial loss.
Such mechanism ensures adversarial examples generated during training are not specific to one architecture.
% which enables us to find architectures that are more generally robust to adv samples.}
We also provide the pseudo code of our robust search algorithm in the Appendix.

% \textred{see supplementary}

% One-shot NAS trains a large supernet for once, which consists of all candidate architectures, and then allow these candidate architectures to share the weights therein later. One appealing advantage of one-shot NAS approach is that once we have trained the supernet, we can easily access to the performance of a large amount of candidate architectures by sampling and inheriting weights from the supernet. The process eliminates the need for training each individual architecture from scratch, which is very time consuming and of high computational cost. This property of one-shot NAS promotes this work for a large variant of architectures to study the architecture's influence on network robustness. 

% The dropping rate is gradually increased over time using a linear schedule. 
% For each intermediate node $x^{(i)}$, the drop rate of each fan-in edge at the end of training is $r^{1/k}$, where $r$ is a hyper-parameter and $k$ is the number of incoming edges to $x^{(i)}$. 

\vspace{.8ex}
\noindent \textbf{Robustness Evaluation.}
Once obtaining the supernet after robust training, we can collect candidate architectures by random sampling from the supernet and inheriting weights. Rather than direct evaluating the network on validation dataset as vanilla NAS methods, we find that finetuning the sampled network with adversarial training for only a few epochs can significantly improve the performance, which is also observed in \cite{NASfinetune}. The intuition behind finetuning is that while the training scheduler tries to inflate the robustness of each architecture, it yet needs to maintain the overall performance of \emph{all} candidate architectures.
% \textred{thus may degrade the robustness of certain architectures during robust search}.
The adversarial accuracy before and after finetuning for 1,000 randomly sampled candidate architectures is illustrated in Fig.~\ref{fig-acc-improve}. It can be clearly seen that the robustness performance has been largely improved.

After finetuning, we evaluate each candidate architecture on validation samples that are adversarially perturbed by the white-box PGD adversary. We regard the adversarial accuracy as the indicator of the network robustness.

\subsection{Analysis of Cell-Based Architectures}

Having set up the robust search framework, we would like to seek for answers for the first question raised in Sec.~\ref{sec-preliminary}, that what kind of architecture patterns is crucial for adversarial robustness. We first conduct analysis of model robustness for cell-based architectures by following a typical setting in NAS methods \cite{zoph2017learning, liu2018darts}, where the architectures between different cells are shared.

\vspace{.8ex}
\noindent \textbf{Statistical Results.} \label{sec-analysis}
In cell-based setting, we adopt robust architecture search on CIFAR-10. We set the number of intermediate nodes for each cell as $N=4$. Recall that we have 2 non-zero operations and 2 input nodes, so the total number of edges in the search space is 14. This results in a search space with the total complexity $(2^2)^{14}-1\approx10^8$, where each architecture parameter is $\alpha\in\{0,1\}^{2\times14}$. For the training of the supernet, we choose 7-step PGD adversarial training with $0.01$ step size. After training the supernet, we randomly sample 1,000 architectures from the supernet and finetune each of them for 3 epochs.  
% $\prod_{m=1}^4\frac{m(m+1)}{2}\times (2^2)^2\approx$

\begin{figure}[tb]
\centering
\subfigure[]{\includegraphics[width=.47\linewidth]{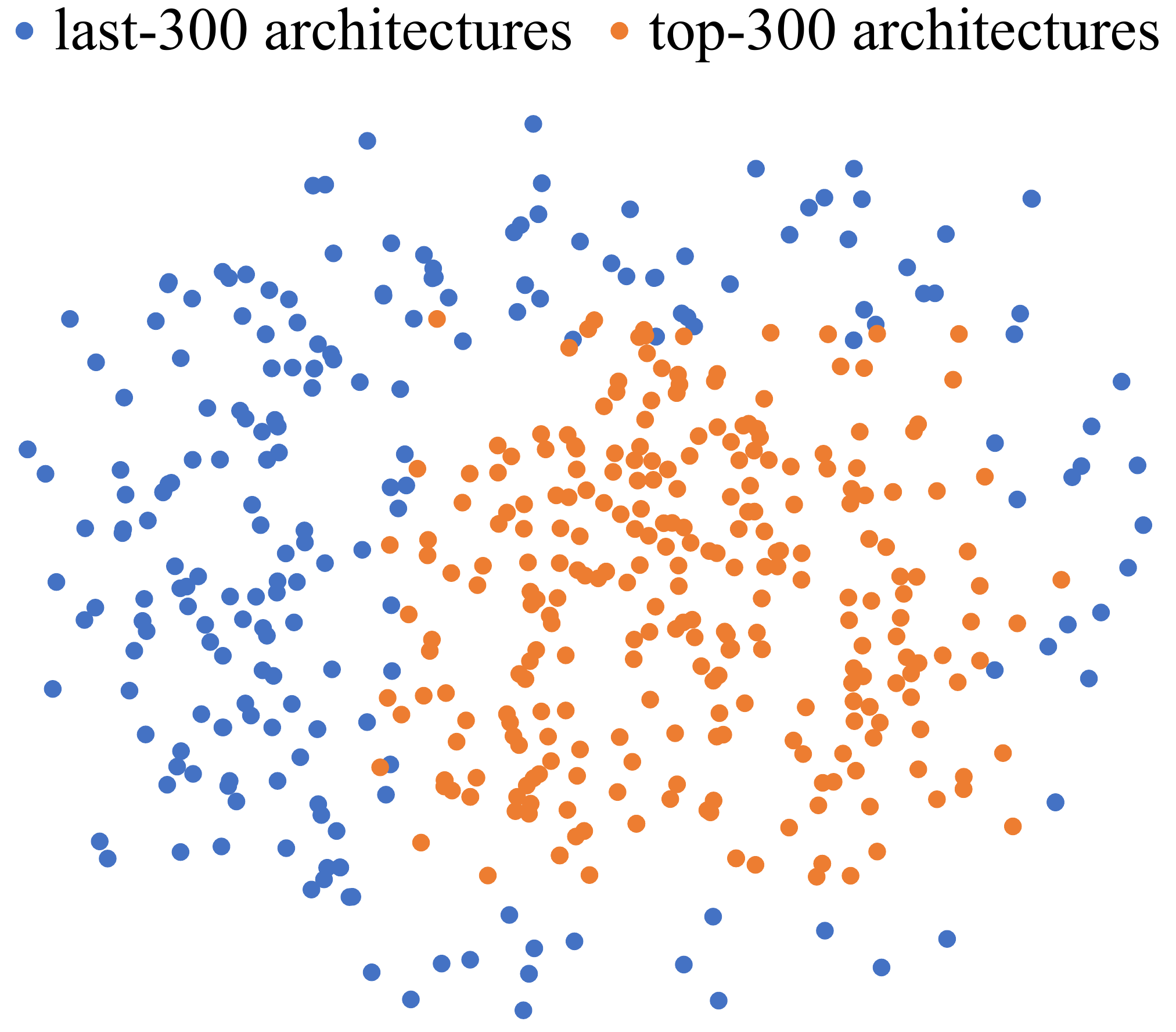}\label{fig-tsne}}
\hspace{1.5ex}
\subfigure[]{\includegraphics[width=.47\linewidth]{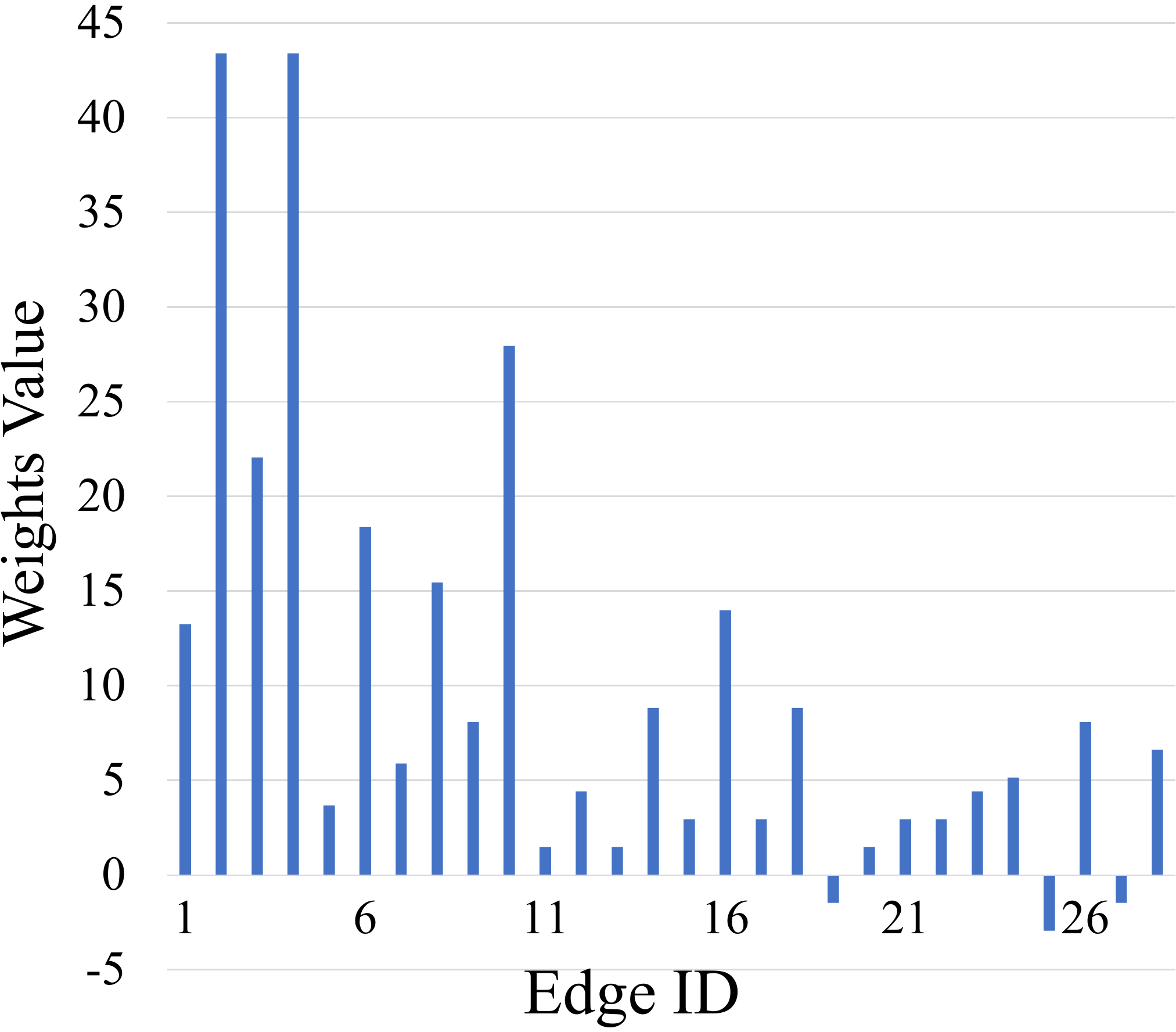}\label{fig-linear}}
\caption{Analysis of the selected top 300 (robust) architectures and last 300 (non-robust) architectures: (a) Visualization of t-SNE on $\alpha$ for all 600 architectures. The embedding of $\alpha$ is separable between robust and non-robust networks, which demonstrates the architecture has an influence on network robustness.
(b) Values of weights of the trained linear classifier. We observe that almost all of the weight values are positive, indicating that there is a strong correlation between architecture density and adversarial accuracy.}
\vspace{-.3cm}
\end{figure}

We plot the histogram of adversarial accuracy of these 1,000 architectures in Fig.~\ref{fig-hist}. As the figure shows, although most of the architectures achieve relatively high robustness (with $\sim50$\% robust accuracy), there also exist a large number of architectures suffering from poor robustness (far lower from the average 50.3\%). This motivates us to consider whether there exist some shared features among the robust networks.
% the robust (non-robust) networks share some common patterns in architectures.

To better visualize how distinguishable the architectures are, we first sort the 1,000 architectures with respect to the robust accuracy. Next, top 300 architectures are selected with a label of $1$ and last 300 architectures with label of $-1$. Finally, t-SNE helps us to depict the $\alpha$ corresponding to each architecture. We visualize the low-dimensional embedding of 600 $\alpha$ in Fig.~\ref{fig-tsne}. As shown, the embedding of architecture parameter $\alpha$ is separable between robust and non-robust networks, which clearly demonstrates that architecture has an influence on network robustness.

\begin{figure}[tb]
\begin{center}
\includegraphics[width=0.95\linewidth]{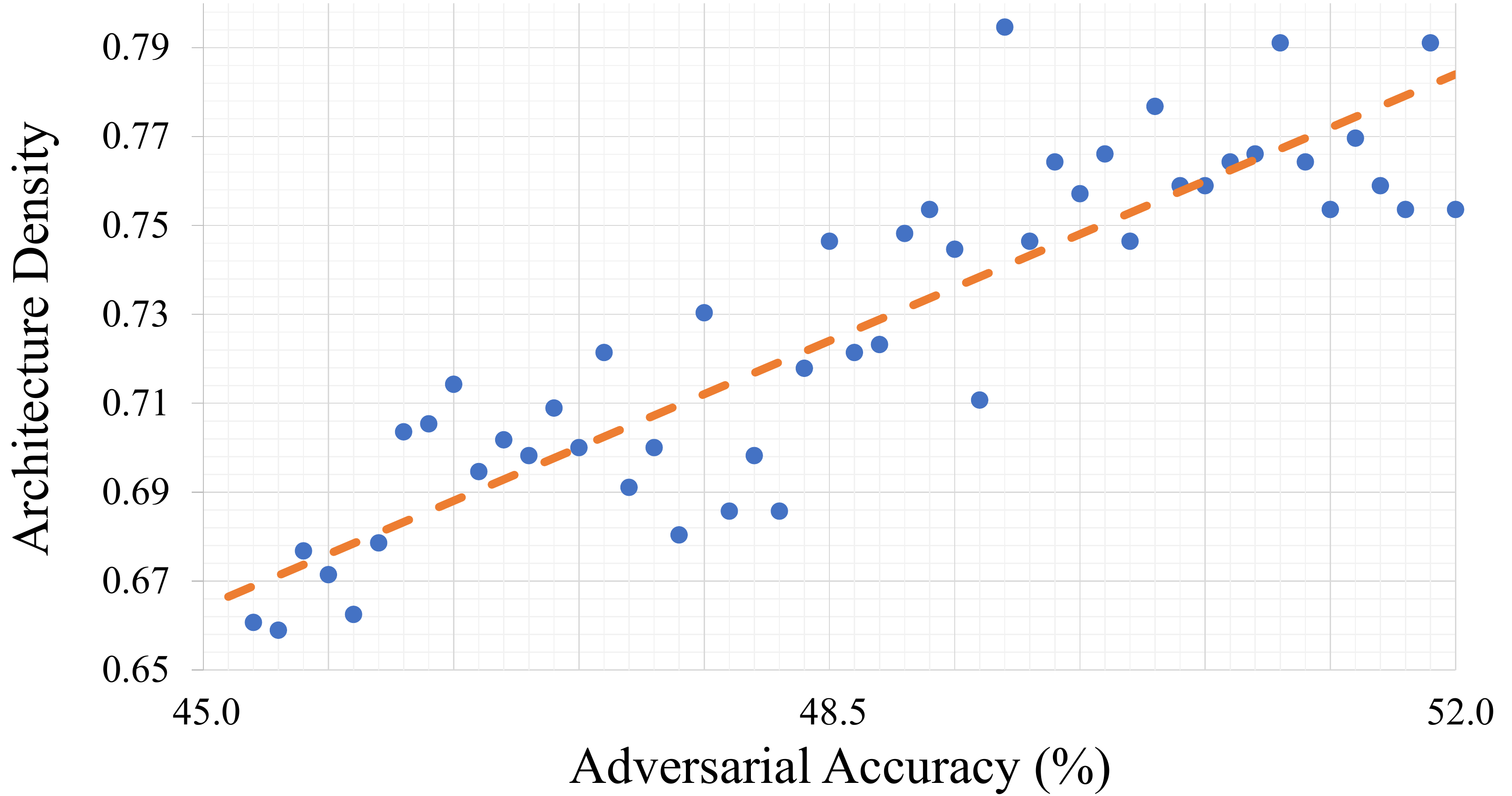}
\end{center}
\vspace{-0.3cm}
\caption{Correlation between architecture density and adversarial accuracy. We show a strong correlation between them, indicating that densely connected pattern can benefit the network robustness.}
\vspace{-0.2cm}
\label{fig-dense}
\end{figure}

This finding naturally raises a question: \emph{Which paths are crucial to network robustness in architectures?} A straightforward idea is that we train a classifier which takes the architecture parameter as input and predicts whether the architecture is robust to adversarial attacks. In this case, the weights that correspond to crucial paths are expected to have larger values. We use the 600 architectures introduced above and their corresponding labels to train a classifier. Surprisingly, we find out that even a \emph{linear} classifier\footnote{\url{https://scikit-learn.org/stable/modules/generated/sklearn.linear_model.SGDClassifier.html}} fits the data well (the training accuracy of these 600 data points is 98.7\%). The results are illustrated in Fig.~\ref{fig-linear}. 
The figure reveals that almost all of the weight values are positive, which indicates a strong relationship between how denser one architecture is wired and how robust it is under adversarial attacks.
% \textred{It is counter-intuitive since a denser network obtains features from different levels, where perturbation is easier to amplified and harm the network robustness.} 
To further explore the relationship, we perform an analysis of the correlation between adversarial accuracy and the density of the architecture. We define the \emph{architecture density} $D$ as the number of connected edges over the total number of all possible edges in the architecture, which can be expressed as:
\begin{eqnarray}
D = \frac{|E_{connected}|}{|E|}=\frac{\sum_{i,j,k}\alpha_{k}^{(i,j)}}{|E|}.
\end{eqnarray}
We illustrate the result in Fig.~\ref{fig-dense}, which shows that there is a strong correlation between architecture density and adversarial accuracy. We posit that through adversarial training, densely connected patterns in the network are more beneficial against adversarial features and learn to be resistant to them. This gives us the answer to the first question in Sec.~\ref{sec-intro}: \emph{Densely connected pattern can benefit the network robustness}.

\subsection{Architecture Strategy under Budget}
% \textred{In this section, we conduct analysis to answer the second question in Sec.~\ref{sec-intro}. }
It has been observed in many previous studies \cite{su2018robustness, madry2017towards} that, within the same family of architectures, increasing the number of parameters of the network would bring improvement of robustness. This is because such procedure will promote the model capacity, and thus can benefit the network robustness. However, if we are given a fixed total number of parameters (or we refer to as a computational budget), how to obtain architectures that are more robust under the limited constraint? In this section, we concentrate on how the pattern of an architecture influences robustness when given different fixed computational budgets. One advantage of our robust architecture search space for this study is that, the number of parameters of a network is positively correlated to the number of convolution operations in the architecture. 

\begin{figure}[tb]
\centering
\subfigure[]{\includegraphics[width=.495\linewidth]{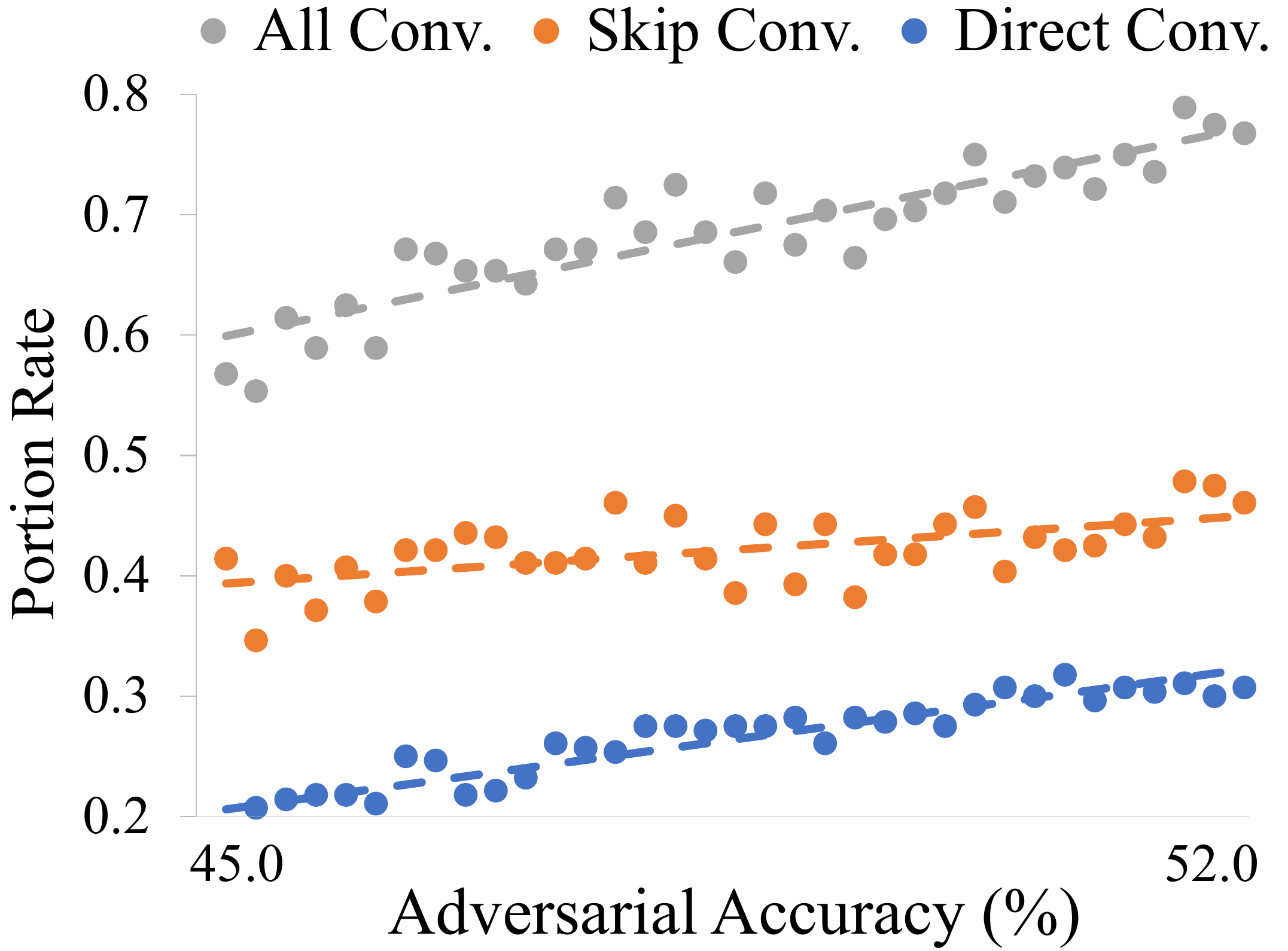}\label{fig-budget-1}}
\subfigure[]{\includegraphics[width=.495\linewidth]{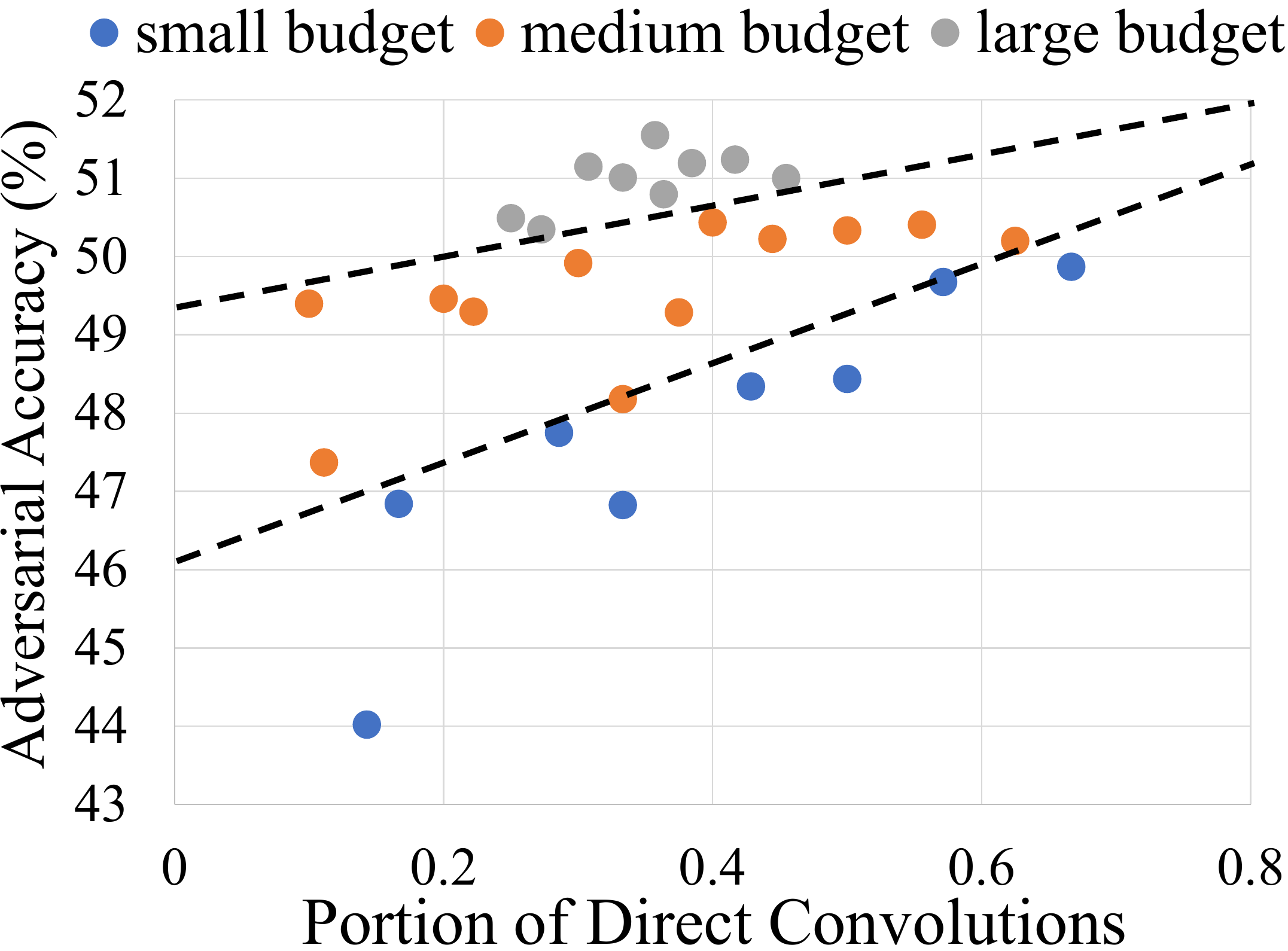}\label{fig-budget-2}}
\caption{Architecture studies under computational budget: (a) Correlation between the number of different operations and the network robustness. When increasing the number of convolution operations, the adversarial accuracy increases steadily. Moreover, convolutions on direct edges contribute more on robust accuracy than those on skip edges. (b) Performance under different computational budgets. Under \emph{small} and \emph{medium} budget, the proportion of direct convolutions shows a positive correlation to adversarial accuracy, indicating that adding convolution operations to direct edges is more effective to improve model robustness under small computational budget.}
\vspace{-.5cm}
\end{figure}

We first analyze the number of convolution operations with respect to network robustness, using the 1,000 architectures obtained in Sec.~\ref{sec-analysis}. The results are illustrated in Fig.~\ref{fig-budget-1}. With the number of convolution operations increases, the adversarial accuracy improves steadily. We also plot the statistics for the number of convolutions on skip edges and direct edges, respectively. The results declare that convolutions on direct edges contribute more on adversarial accuracy than those on skip edges. This inspires us to dig deeper on the effect of the convolutions on direct edges for different computational budgets.

We consider three different computational budgets: \emph{small}, \emph{medium} and \emph{large}. Since the maximum number of convolution operations in the cell-based setting is 14, we set the total number of convolutions smaller than 7 as \emph{small} budget, between 8 and 10 as \emph{medium} and larger than 11 as \emph{large}. For each of the budget, we randomly sample 100 architectures, evaluate their adversarial accuracy following Sec.~\ref{search-algorithm} and calculate the proportion of convolutions on direct edges among all convolutions. As illustrated in Fig.~\ref{fig-budget-2}, the adversarial accuracy has clear boundaries between different budgets. Furthermore, for \emph{small} and \emph{medium} budget, the proportion of direct convolutions has a positive correlation to adversarial accuracy. This indicates that for smaller computational budget, adding convolutions to direct edges can efficiently improve the network robustness. We also note that this phenomenon is not obvious for the \emph{large} setting. We speculate that for architectures within the \emph{large} budget, densely connected patterns dominate the contributions of network robustness. With the above results, we conclude: \emph{Under small computational budget, adding convolution operations to direct edges is more effective to improve model robustness.}

\subsection{Towards a Larger Search Space}\label{sec:fsp guided search}

\noindent \textbf{Relax the Cell-Based Constraint.}
In previous sections, we obtain several valuable observations for the cell-based setting. One natural question to ask is: What if we relax the constraint and permit all the cells in the network to have different architectures? Moreover, what can be the indicator for the network robustness in this cell-free setting? In this section, we relax the cell-based constraint and conduct studies on a larger architecture search space. The relaxation of the constraint raises an explosion of the complexity of the search space: for a network consisting of $L$ cells, the total complexity increases to $(10^8)^L$. The exponential complexity makes the architecture search much more difficult to proceed.

\vspace{.8ex}
\noindent \textbf{Feature Flow Guided Search.}
To address the above challenges, here we propose a feature flow guided search scheme. Our approach is inspired by TRADES~\cite{zhang2019theoretically}, which involves a loss function minimizing the KL divergence of the logit distribution between an adversarial example and its corresponding clean data. The value of this loss function can be utilized as a measurement of the gap between network robustness and its clean accuracy. Instead of focusing on the final output of a network, we consider the feature flow between the intermediate cells of a network. Specifically, we calculate the Gramian Matrix across each cell, denoted as flow of solution procedure (FSP) matrix \cite{yim2017gift}. The FSP matrix for the $l$th cell is calculated as:
\begin{eqnarray}
G_l(x;\theta) = \sum_{s=1}^{h}\sum_{t=1}^{w}\frac{F_{l,s,t}^{in}(x;\theta)\times F_{l,s,t}^{out}(x;\theta)}{h\times w},
\end{eqnarray}
where $F_l^{in}(x;\theta)$ denotes the input feature map of the cell and $F_l^{out}(x;\theta)$ denotes the output feature map. For a given network, we can calculate the distance of FSP matrix between adversarial example and clean data for each cell of the network:
\begin{eqnarray}
L^{FSP}_l = \frac{1}{N} \sum_x\left\|(G_{l}(x;\theta) - G_{l}(x';\theta)\right\|_2^2.
\end{eqnarray}

We sample 50 architectures $without$ finetuning for the cell-free search space and evaluate the gap of clean accuracy and adversarial accuracy for each architecture. We also calculate the FSP matrix distance for each cell of the network and show representative results in Fig.~\ref{fig-gram-1} (complete results are provided in Appendix). We can observe that for the cells in a deeper position of the network, the FSP distance has a positive correlation with the gap between network robustness and clean accuracy. This gives us the answer to the third question in Sec.~\ref{sec-intro}: \emph{A robust network has a lower FSP matrix loss in the deeper cells of network.}

By observing this phenomenon, we can easily adopt FSP matrix loss to filter out the non-robust architectures with high loss values, which efficiently reduces the complexity of the search space. Thus, after the sampling process from supernet in cell-free setting, we first calculate FSP matrix loss for each architecture and reject those with high loss values. We then perform finetuning to get final robustness.

\begin{figure}[tb]
\centering
\includegraphics[width=1\linewidth]{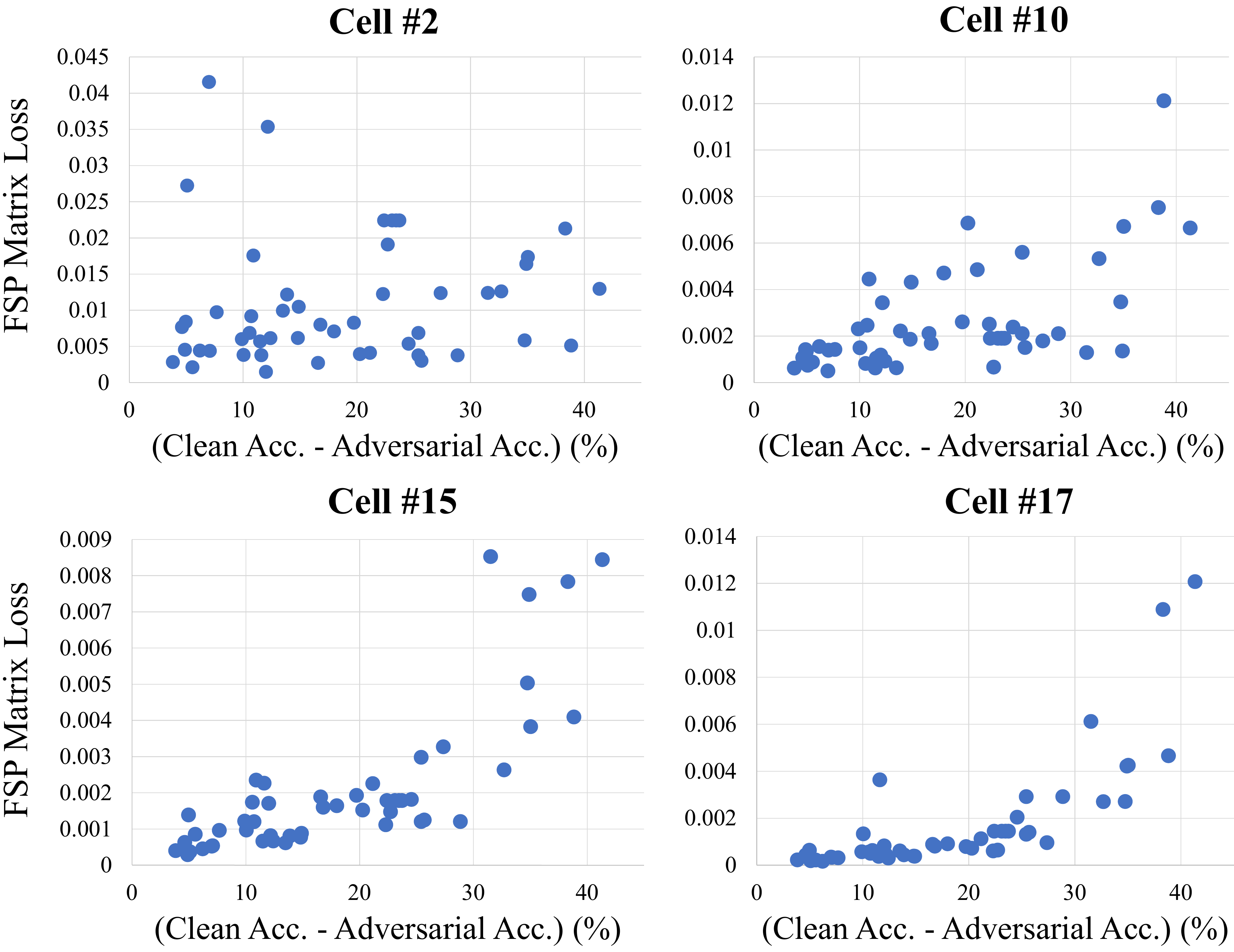}
\caption{Analysis of FSP matrix distance as robustness indicator. We compute the FSP matrix distance for each cell, along with the performance gap between clean accuracy and adversarial accuracy (complete results in Appendix). For cells in deeper positions of the network, the FSP distance has a positive correlation with the gap between network robustness and its clean accuracy, which indicates that a robust network has a lower FSP matrix loss in the deeper cells of the network.}
\vspace{-.3cm}
\label{fig-gram-1}
\end{figure}

\section{Experiments}
In this section, we empirically evaluate the adversarial robustness of the proposed RobNet family.
Following the guidance of our three findings in Sec.~\ref{sec:robust-arch-search}, we train and select a set of representative RobNet models for evaluation.
We focus on $l_\infty$-bounded attacks and compare the RobNet family with state-of-the-art human-designed models.

\begin{table*}[!t]
\caption{White-box attack results on CIFAR-10. We compare representative RobNet models with state-of-the-art architectures. All models are adversarially trained using PGD with 7 steps. All attacks are $l_{\infty}$-bounded with a total perturbation scale of $8/255$~(0.031).}
\vspace{.5ex}
\small
\centering
\scalebox{0.95}{
\setlength{\tabcolsep}{3mm}{
\begin{tabular}{@{}c|c|cccccc@{}}
\toprule
\textbf{Models}               & \textbf{Model Size} & \textbf{Natural Acc.} & \textbf{FGSM}    & \textbf{PGD}$^{20}$ & \textbf{PGD}$^{100}$ & \textbf{DeepFool} & \textbf{MI-FGSM} \\ \midrule\midrule
ResNet-18                     &      11.17M         & 78.38\%               & 49.81\%          & 45.60\%          & 45.10\%         & 47.64\%           & 45.23\%          \\ \midrule
ResNet-50                     &      23.52M               & 79.15\%               & 51.46\%          & 45.84\%          & 45.35\%         & 49.18\%           & 45.53\%          \\ \midrule
WideResNet-28-10              &      36.48M        & \textbf{86.43\%}      & 53.57\%          & 47.10\%          & 46.90\%         & 51.23\%           & 47.04\%          \\ \midrule
DenseNet-121                  &      6.95M               & 82.72\%               & 54.14\%          & 47.93\%          &   47.46\%   & 51.70\%           & 48.19\%          \\ \midrule\midrule
\textbf{RobNet-small}    &      4.41M          & 78.05\%               & {53.93\%} & {48.32\%} & {48.07\%}& {52.96\%}  & {48.98\%} \\ \midrule
\textbf{RobNet-medium}   &      5.66M          & 78.33\%               & {54.55\%} & {49.13\%} & {48.96\%}& {53.32\%}  & {49.34\%} \\ \midrule
\textbf{RobNet-large}    &      6.89M          & 78.57\%               & {54.98\%} & {49.44\%} & {49.24\%}& {53.85\%}  & {49.92\%} \\ \midrule
\textbf{RobNet-large-v2} &     33.42M          & \textbf{85.69\%}      & \underline{57.18\%} & \underline{50.53\%} & \underline{50.26\%}& \underline{55.45\%}  & \underline{50.87\%} \\ \midrule\midrule
\textbf{RobNet-free}          &   5.49M            & 82.79\%               & \textbf{58.38\%} & \textbf{52.74\%} & \textbf{52.57\%}& \textbf{57.24\%}  & \textbf{52.95\%} \\ \bottomrule
\end{tabular}
}}
\vspace{-0.3cm}
\label{table:cifar white}
\end{table*}

\subsection{Experimental Setup}

\noindent {\bf Implementation Details.}
As described in Sec.~\ref{sec:robust-arch-search}, we use both cell-based and cell-free searching algorithm to select out a set of RobNet architectures, respectively.
The robust search is performed only on CIFAR-10, where we use PGD adversarial training with $7$ attack iterations and a step size of $2/255$ (0.01).
For evaluation on other datasets, we directly transfer the RobNet architectures searched on CIFAR-10.

Specifically, we first follow the cell-based robust search framework to obtain architectures that exhibit \emph{densely connected patterns}. Considering the strategy under budget, we further generate three cell-based architectures that all follow \emph{more convolution operations on direct edges}, but with different computational budgets. We refer to the three selected architectures as \textbf{RobNet-small}, \textbf{RobNet-medium}, and \textbf{RobNet-large}.
Furthermore, we leverage \emph{FSP guided search} described in Sec.~\ref{sec:fsp guided search} to efficiently generate cell-free robust architectures and select one representative model for evaluation, which is referred to as \textbf{RobNet-free}.
Note that we are not selecting the \emph{best} architecture, as the searching space is too large to allow us to do so. Instead, we follow the proposed algorithm to select \emph{representative} architectures and study their robustness under adversarial attacks. More details of the selecting process and visualizations of the representative RobNet architectures can be found in Appendix.

% \vspace{.8ex}
% \noindent \textbf{Comparing Architectures.}
%
We compare RobNet with widely used human-designed architectures, including ResNet~\cite{He_2016_CVPR}, Wide-ResNet~\cite{zagoruyko2016wide}, and DenseNet~\cite{huang2017densely}. All models are adversarially trained using PGD with $7$ attack steps and a step size of $2/255$~(0.01).
We follow the training procedure as in~\cite{madry2017towards} and keep other hyper-parameters the same for all models. % during adversarial training.

% \vspace{.8ex}
% \noindent \textbf{Adversarial Attacks.}
%
% As common in prior work~\cite{madry2017towards,song2018pixeldefend,yang2019menet,zhang2019theoretically}, adversarial perturbation is considered under $l_\infty$ norm in this paper. We generate adversarial examples using standard methods such as the Fast Gradient Sign Method (FGSM)~\cite{goodfellow2015explaining}, the C\&W attack~\cite{carlini2017towards}, DeepFool~\cite{moosavi2016deepfool}, Momentum Iterative FGSM (MI-FGSM)~\cite{dong2017boosting}, and Projected Gradient Decent (PGD)~\cite{madry2017towards}.

\vspace{.8ex}
\noindent {\bf Datasets \& Evaluation.}
We first perform an extensive study on CIFAR-10 to validate the effectiveness of RobNet against black-box and white-box attacks.
We then extend the results to other datasets such as SVHN, CIFAR-100, and Tiny-ImageNet.
Finally, we show the benefits from RobNet are orthogonal to existing techniques.
% d show further improvements.
We provide additional results on ImageNet, as well as detailed training procedure and hyper-parameters in Appendix.

\subsection{White-box Attacks}
\label{sec:white-box experiment}
% In white-box attacks, the adversary has full access to the model architecture and weights.
% We show the effectiveness of our searched robust architectures on top of adversarial training~\cite{madry2017towards,goodfellow2015explaining}. Following the recent success on PGD adversarial training~\cite{madry2017towards}, we train all models using PGD with 7 attack steps and 

% \vspace{.8ex}
\noindent {\bf Main Results.}
We show the results against various white-box attacks in Table~\ref{table:cifar white}. We choose state-of-the-art network architectures that are widely used in adversarial literature \cite{madry2017towards,xie2019feature,zhang2019theoretically} for comparison.
As illustrated in the table, all the selected models from RobNet family can consistently achieve higher robust accuracy under different white-box attacks, compared to other models. 
% Moreover, \textred{the model sizes of RobNet are smaller than other prevailing network architectures.}

The strongest adversary in our white-box setting is the PGD attacker with 100 attack iterations~(i.e., PGD$^{100}$). When zoom in to the results, we can observe that by only changing architecture, RobNet can improve the previous arts under white-box attacks by \textbf{5.1\%} from 47.5\% to 52.6\%.

\vspace{.8ex}
\noindent {\bf The Effect of Dense Connections.}
Table~\ref{table:cifar white} also reveals interesting yet important findings on dense connections. ResNet and its wide version~(WideResNet) are most frequently used architectures in adversarial training~\cite{madry2017towards,xie2019feature,zhang2019theoretically}. % where increasing capacity alone~(e.g., WideResNet \emph{vs.} ResNet) is reported to be help increasing robustness~\cite{madry2017towards}.
Interestingly however, it turns out that the rarely used DenseNet model is more robust than WideResNet, even with much fewer parameters. Such observation are well-aligned with our previous study: \emph{densely connected pattern} largely benefits the model robustness.
Since RobNet family explicitly reveals such patterns during robust architecture search, they turn out to be consistently robust.

\vspace{.8ex}
\noindent {\bf The Effect of Parameter Numbers.}
Inspired by the finding of computational budget, we seek to quantify the robustness of RobNets with different parameter numbers. We compare three models with different sizes obtained by cell-based search (i.e., RobNet-small, RobNet-medium, and RobNet-large). As Table~\ref{table:cifar white} reports, with larger computational budgets, network robustness is consistently higher, which is well aligned with our arguments.

We note that the model sizes of RobNets are consistently smaller than other widely adopted network architectures.
Yet, the natural accuracy of RobNet model is unsatisfying when compared to WideResNet.
To further study the influence of network parameters, we extend the RobNet-large model to have similar size as WideResNet by increasing the number of channels and stacked cells, while maintaining the same architecture within each cell. We refer to this new model as \textbf{RobNet-large-v2}.
% Moreover, recent work~\cite{madry2017towards} argues that by increasing the model capacity alone can help improve the robustness of network. This can be verified by comparing the results between ResNet and WideResNet.
% As restricted by the computational budget, all RobNet-CB models have much smaller size than WideResNet, then how would robustness change when scaling up the model size remains to be interesting.
% As Table~\ref{table:cifar white} 
It turns out that by increasing the model size, not only can the robustness be strengthened, the natural accuracy can also be significantly improved.

\vspace{.8ex}
\noindent {\bf The Effect of Feature Flow Guided Search.}
When releasing the cell-based constraints during robust searching, RobNet models can be even more robust. We confirm it by comparing the results of RobNet-free, which is obtained using FSP Guided Search as mentioned in Sec.~\ref{sec:fsp guided search}, with other cell-based RobNet models.
Remarkably, RobNet-free achieves higher robust accuracy with 6$\times$ fewer parameter numbers when compared to RobNet-large-v2 model.

\begin{table}[!t]
\caption{Black-box attack results on CIFAR-10. We compare two representative RobNet architectures with state-of-the-art models. %All models are adversarially trained.
Adversarial examples are generated using transfer-based attack on the same copy of the victim network.}
\vspace{1ex}
\centering
\small
\scalebox{0.9}{
\setlength{\tabcolsep}{4.5mm}{
\begin{tabular}{@{}c|cc@{}}
\toprule
\textbf{Models}      & \textbf{FGSM}    & \textbf{PGD}$^{100}$\\ \midrule\midrule
ResNet-18            & 56.29\%          & 54.28\%          \\ \midrule
ResNet-50            & 58.12\%          & 55.89\%          \\ \midrule
WideResNet-28-10     & 58.11\%          & 55.68\%          \\ \midrule
DenseNet-121         & 61.87\%          & 59.34\%          \\ \midrule\midrule
\textbf{RobNet-large}& \textbf{61.92\%} & \textbf{59.58\%} \\ \midrule
\textbf{RobNet-free} & \textbf{65.06\%} & \textbf{63.17\%} \\ \bottomrule
\end{tabular}
}}
% \vspace{-.3cm}
\label{table:cifar black}
\end{table}

\subsection{Black-box Attacks}
We further verify the robustness of RobNet family under black-box attacks. We follow common settings in literature \cite{papernot2016transferability,madry2017towards,yang2019menet} and apply transfer-based black-box attacks. We train a copy of the victim network using the same training settings, and apply FGSM and PGD$^{100}$ attacks on the copy network to generate adversarial examples. Note that we only consider the \emph{strongest} transfer-based attacks, i.e., we use \emph{white-box attacks} on the independently trained copy to generate black-box examples.

The results are shown in Table~\ref{table:cifar black}. They reveal that both cell-based and cell-free RobNet models are more robust under transfer-based attacks. Note that here the source model has the same architecture as the target model, which makes the black-box adversary stronger~\cite{madry2017towards}. We also study the transfer between different architectures, and provide corresponding results in the Appendix.

\begin{table}[!t]
\caption{White-box attack results across different datasets. We use two RobNet models searched on CIFAR-10 to directly perform adversarial training on new datasets. We apply PGD$^{100}$ white-box attack on all models to evaluate adversarial robustness.}
\vspace{1ex}
\centering
\small
\scalebox{0.9}{
\setlength{\tabcolsep}{2mm}{
\begin{tabular}{@{}c|ccc@{}}
\toprule
\textbf{Models}      & \textbf{SVHN} & \textbf{CIFAR-100} & \textbf{Tiny-ImageNet}\\ \midrule\midrule
ResNet-18            & 46.08\%    &   22.01\% &     16.96\%     \\ \midrule
ResNet-50            & 47.23\%    &   22.38\% &     19.12\%     \\ \midrule\midrule
\textbf{RobNet-large}& \textbf{51.26\%} & \textbf{23.19\%} & \textbf{19.90\%} \\ \midrule
\textbf{RobNet-free} & \textbf{55.59\%} & \textbf{23.87\%} & \textbf{20.87\%} \\ \bottomrule
\end{tabular}
}}
\vspace{-.3cm}
\label{table:other datasets}
\end{table}

\subsection{Transferability to More Datasets}
So far, our robust searching has only been performed and evaluated on CIFAR-10. However, the idea of robust neural architecture search is much more powerful: we directly use the RobNet family searched on CIFAR-10 to apply on other datasets, and demonstrate their effectiveness. Such benefits come from the natural advantage of NAS that the searched architectures can generalize to other datasets~\cite{zoph2017learning, zhong2018practical}.

We evaluate RobNet on SVHN, CIFAR-100, and Tiny-ImageNet under white-box attacks, and set attack parameters as follows: total perturbation of $8/255$ (0.031), step size of $2/255$ (0.01), and with 100 total attack iterations. The training procedure is similar to that on CIFAR-10, where we use 7 steps PGD for adversarial training. We keep all the training hyperparameters the same for all models.

Table~\ref{table:other datasets} shows the performance of RobNet on the three datasets and compares them with commonly used architectures.
The table reveals the following results. First, it verifies the effectiveness of RobNet family: they consistently outperform other baselines under strong white-box attacks.
Furthermore, the gains across different datasets are different. RobNets provide about 2\% gain on CIFAR-100 and Tiny-ImageNet, and yield $\sim$10\% gain on SVHN.

\begin{table}[!t]
\caption{Robustness comparison of different architectures with and without feature denoising \cite{xie2019feature}. We show the benefits from RobNet are orthogonal to existing techniques: RobNet can further boost robustness performance when combined with feature denoising.}
\vspace{1ex}
\centering
\small
\scalebox{0.9}{
\begin{tabular}{@{}c|ccc@{}}
\toprule
\textbf{Models}        & \textbf{Natural Acc.} & \textbf{PGD$^{100}$} \\ \midrule\midrule
ResNet-18              & 78.38\%               & 45.10\%         \\ \midrule
ResNet-18 + Denoise    & 78.75\%               & 45.82\%         \\ \midrule\midrule
\textbf{RobNet-large}           & 78.57\%               & 49.24\%         \\ \midrule
\textbf{RobNet-large + Denoise} & \textbf{84.03\%}      & \textbf{49.97\%}\\ \bottomrule
\end{tabular}
}
\vspace{-.3cm}
\label{table:feature denoise}
\end{table}

\subsection{Boosting Existing Techniques}
As RobNet improves model robustness from the aspect of network architecture, it can be seamlessly incorporated with existing techniques to further boost adversarial robustness.
To verify this advantage, we select feature denoising technique \cite{xie2019feature} which operates by adding several denoising blocks in the network.
We report the results in Table~\ref{table:feature denoise}. As shown, the denoising module improves both clean and robust accuracy of RobNet, showing their complementariness.
Moreover, when compared to ResNet-18, RobNet can better harness the power of feature denoising, gaining a larger improvement gap, especially on clean accuracy.
% complementary

\section{Conclusion}

We proposed a robust architecture search framework, which leverages one-shot NAS to understand the influence of network architectures against adversarial attacks. Our study revealed several valuable observations on designing robust network architectures. Based on the observations, we discovered RobNet, a family of robust architectures that are resistant to attacks. Extensive experiments validated the significance of RobNet, yielding the intrinsic effect of architectures on network resilience to adversarial attacks.
% We esteem our findings to deepen the understanding towards the intriguing property of adversarial examples, and future work exploring adversarial attack may pay more attention on the intrinsic effect of architectures.

{\small
\bibliographystyle{ieee_fullname}
\bibliography{egbib}
}
\clearpage

\begin{appendices}
%%%%%%%%% BODY TEXT
\section{Details of Robust Architecture Search}
We provide details of our robust architecture search algorithm. The pseudo code is illustrated in Algorithm \ref{alg-robust-search}.

\begin{algorithm}[ht]
\caption{Robust architecture search}
\begin{algorithmic}[1]
\STATE {\bfseries Input:} Supernet $G=(V,E)$, architecture parameter $\alpha$, total iterations $I$, PGD attack iterations $T$.
% the set of identity edges $e=\{e^{(i,j)}\}$, the adversarial accuracy $P_0$ and threshold $\sigma$;}
\STATE {Set all elements in $\alpha$ to $1$}
\FOR{$k=0\dots I$}
\STATE {Randomly sample a training batch $\{x_i,y_i\}_{i=1}^B$ from train dataset}
\STATE {Randomly set some of the elements in $\alpha$ to 0 and get the corresponding network parameter $\theta_k$}
\STATE \texttt{/*}~~\texttt{Parallel}~~\texttt{training}~~\texttt{in}~~\texttt{PyTorch}~\texttt{*/}
\FOR{$i=1\dots B$}
\STATE $x^{(0)}_i \leftarrow x_i$
\STATE \texttt{/*}~~\texttt{PGD}~~\texttt{adversarial}~~\texttt{example}~\texttt{*/}
\FOR{$t=0\dots (T-1)$}
\STATE {$x^{(t+1)}_i \leftarrow \Pi_{S}(x^{(t)}_i + \eta\cdot {\rm{sign}}(\nabla_{x}\mathcal{L}(\theta_k,x^{(t)}_i,y_i))$}
% \STATE {clip}
\ENDFOR
\ENDFOR
\STATE Use $\{x^{(T)}_i,y_i\}_{i=1}^B$ to do one step training and update $\theta_k$ by SGD
\STATE Set all elements in $\alpha$ to $1$
\ENDFOR
\end{algorithmic}
\label{alg-robust-search}
% \vspace{-.1cm}
\end{algorithm}

\section{Details of Adversarial Training}
\label{appendix:training details}
We further provide training details of PGD-based adversarial training for each individual architecture on CIFAR, SVHN, and Tiny-ImageNet.
We summarize our training hyper-parameters in Table~\ref{table:appendix training}. We follow the standard data augmentation scheme as in~\cite{He_2016_CVPR} to do zero-padding with 4 pixels on each side, and then random crop back to the original image size. We then randomly flip the images horizontally and normalize them into $[0,1]$.
We use the same training settings for CIFAR-10 and CIFAR-100.

% =========================================================================== %
\begin{table}[!t]
\caption{Details of adversarial training on different datasets. Learning rate is decreased at selected epochs, using a step factor of 0.1. We apply the same training setting for both CIFAR-10 and CIFAR-100.}
\vspace{1ex}
\centering
\scalebox{0.95}{
\setlength{\tabcolsep}{2.5mm}{
\begin{tabular}{@{}c|ccc@{}}
\toprule
                   & \textbf{CIFAR}  & \textbf{SVHN}   & \textbf{Tiny-ImageNet} \\ \midrule\midrule
\textbf{Optimizer} & SGD             & SGD             & SGD                    \\\midrule
\textbf{Momentum}  & 0.9             & 0.9             & 0.9                    \\\midrule
\textbf{Epochs}    & 200             & 200             & 90                     \\\midrule
\textbf{LR}        & 0.1             & 0.01            & 0.1                    \\\midrule
\textbf{LR decay}  & \begin{tabular}[c]{@{}c@{}}step\\ (100, 150)\end{tabular} & \begin{tabular}[c]{@{}c@{}}step\\ (100, 150)\end{tabular} & \begin{tabular}[c]{@{}c@{}}step\\ (30, 60)\end{tabular} \\ \bottomrule
\end{tabular}
}}
\vspace{-1ex}
\label{table:appendix training}
\end{table}
% =========================================================================== %

\section{Additional Results on ImageNet}
In this section, we provide additional robustness results of RobNets on ImageNet \cite{deng2009imagenet}, a large-scale image classification dataset that contains $\sim$1.28 million images and 1000 classes.
Since adversarial training on ImageNet demands a vast amount of computing resources (e.g., hundreds of GPUs \cite{xie2019feature,kannan2018adversarial} for several days), we adopt the recent ``free'' adversarial training scheme \cite{shafahi2019adversarial} for accelerating adversarial training on ImageNet.
Specifically, we follow the settings in \cite{shafahi2019adversarial} that consider non-targeted attack, and restrict the perturbation bound to be $\epsilon=4/255$~(0.015). We use $m=4$ for the ``free'' training, and keep other hyper-parameters the same for all models.

We compare RobNet-large model with different variants of ResNet against white-box PGD attacks. The results are shown in Table~\ref{table:appendix imagenet}. Compared to different ResNet models, RobNet-large can consistently achieve higher robust accuracy against PGD adversary, while maintaining similar clean accuracy. We note that the model size of RobNet-large is far smaller than the baseline models. As we have investigated in Sec.~\ref{sec:white-box experiment}, by increasing the network parameters of RobNet models, we can not only strengthen the adversarial robustness, the natural accuracy can also be significantly improved.
This phenomenon can also be observed by comparing ResNet models with different capacities.
Thus, we believe by further increasing the parameter numbers, RobNets can achieve even higher accuracy in both clean and adversarial settings.
% RobNet-large achieves \emph{40.1\%} compared to ResNet-50 with \emph{36.4\%}, under PGD$^{50}$ white-box attack within $\epsilon=4/255$.

% =========================================================================== %
\begin{table*}[!t]
\caption{White-box attack results on ImageNet. We compare representative RobNet models with state-of-the-art architectures. All models are adversarially trained using ``free'' training \cite{shafahi2019adversarial}. All attacks are $l_{\infty}$-bounded with a total perturbation scale of $4/255$~(0.015).}
\vspace{1ex}
\centering
\scalebox{0.95}{
\setlength{\tabcolsep}{3mm}{
\begin{tabular}{@{}c|c|cccc@{}}
\toprule
\textbf{Models} & \textbf{Model Size} & \textbf{Natural Acc.} & \textbf{PGD}$^{10}$ & \textbf{PGD}$^{50}$ & \textbf{PGD}$^{100}$ \\ \midrule\midrule
ResNet-50       &  23.52M   & 60.20\%               & 32.76\%          & 31.87\%          & 31.81\% \\ \midrule
ResNet-101      &  42.52M   & 63.34\%               & 35.38\%          & 34.40\%          & 34.32\% \\ \midrule
ResNet-152      &  58.16M   & \textbf{64.44\%}      & 36.99\%          & 36.04\%          & 35.99\% \\ \midrule\midrule
\textbf{RobNet-large}  &   12.76M   & 61.26\%        & \textbf{37.16\%} & \textbf{37.15\%} & \textbf{37.14\%} \\ \bottomrule
\end{tabular}
}}
% \vspace{-0.3cm}
\label{table:appendix imagenet}
\end{table*}
% =========================================================================== %

\section{Comparisons to More Architectures}
In this section, we provide a more comprehensive comparison between RobNet models and various state-of-the-art human-designed architectures. In addition to ResNet and DenseNet family we have mentioned in the main text, we further add baseline architectures including VGG \cite{simonyan2014very}, MobileNetV2 \cite{sandler2018mobilenetv2}, and ResNeXt \cite{xie2017aggregated}, and report the results in Table \ref{table:appendix more architects}.
We again consider $l_{\infty}$-bounded white-box attack setting on CIFAR-10, with all models trained identically as we have described. 
As can be observed from the table, when comparing to various human-designed architectures, RobNet models can consistently achieve higher adversarial robustness, even with much fewer network parameters.

% =========================================================================== %
\begin{table}[!t]
\caption{Comparison between representative RobNet models and more human-designed architectures on CIFAR-10. All models are adversarially trained using PGD with 7 steps. All attacks are $l_{\infty}$-bounded with a total perturbation scale of $8/255$~(0.031).}
\vspace{1ex}
\centering
\scalebox{0.91}{
\begin{tabular}{@{}c|c|c|c@{}}
\toprule
\textbf{Models} & \textbf{Model Size} & \textbf{Natural Acc.} & \textbf{PGD}$^{100}$ \\ \midrule\midrule
VGG-16  &  14.73M  &  77.38\%  &  44.38\%  \\ \midrule
ResNet-18  &  11.17M  &  78.38\%  &  45.10\%  \\ \midrule
ResNet-50  &  23.52M  &  79.15\%  &  45.35\%  \\ \midrule
MobileNetV2  &  2.30M  &  76.79\%  &  45.50\%  \\ \midrule
ResNeXt-29 (2x64d)  &  9.13M  & 81.86\%  &  46.04\%  \\ \midrule
WideResNet-28-10  &  36.48M  &  \textbf{86.43\%}  &  46.90\%  \\ \midrule
DenseNet-121  &  6.95M  & 82.72\%  &  47.46\%  \\ \midrule\midrule
\textbf{RobNet-small}    &  4.41M  &  78.05\%  &  48.07\%  \\ \midrule
\textbf{RobNet-medium}   &  5.66M  &  78.33\%  &  48.96\%  \\ \midrule
\textbf{RobNet-large}    &  6.89M  &  78.57\%  &  49.24\%  \\ \midrule
\textbf{RobNet-large-v2} &  33.42M &  \textbf{85.69\%}  &  \underline{50.26\%}  \\ \midrule\midrule
\textbf{RobNet-free}     &  5.49M  &  \underline{82.79\%}  &  \textbf{52.57\%}  \\ \bottomrule
\end{tabular}
}
% \vspace{-0.3cm}
\label{table:appendix more architects}
\end{table}
% =========================================================================== %

\section{Complete Results of FSP Matrix Loss}
\begin{figure*}[tb]
\centering
\includegraphics[width=.99\linewidth]{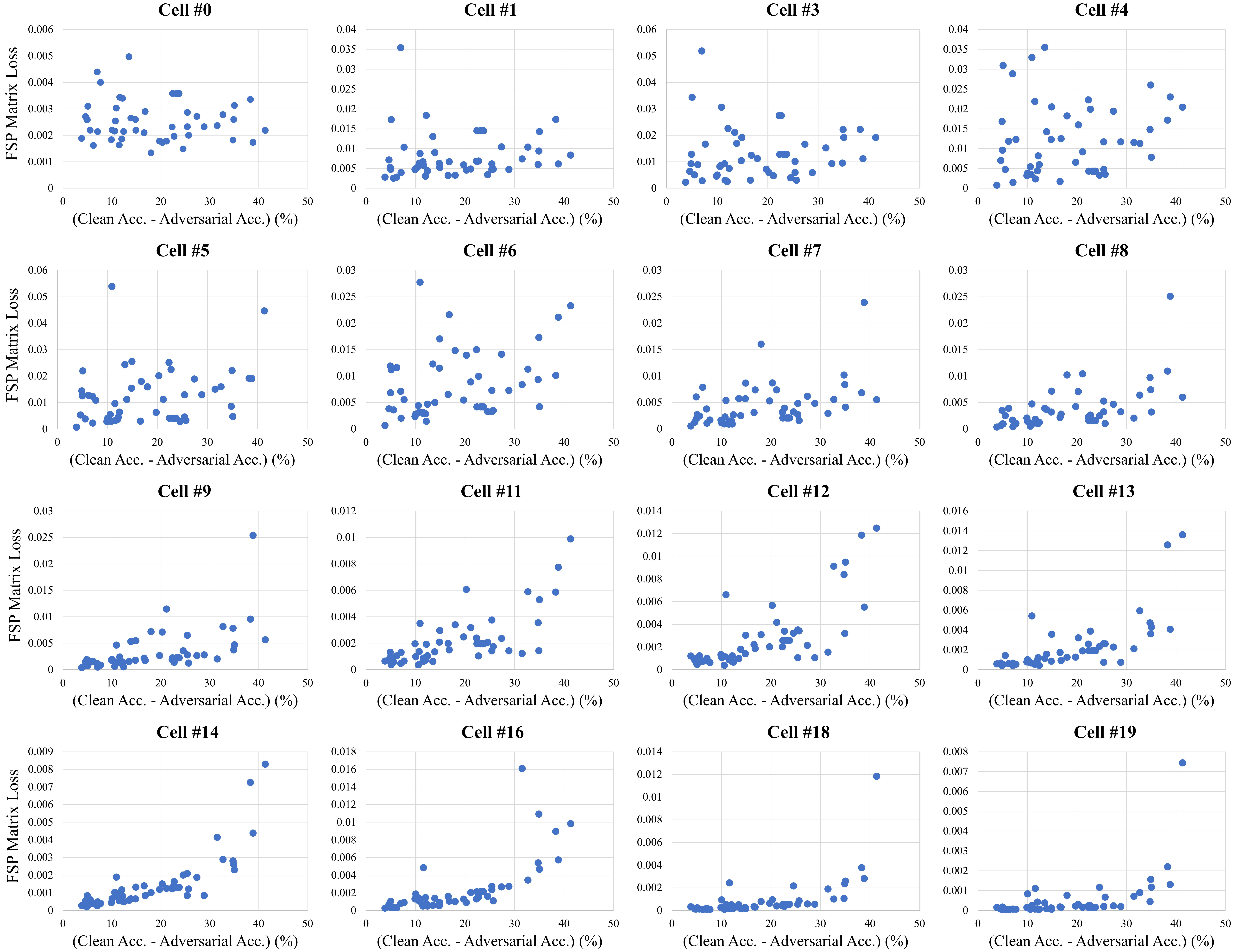}
\vspace{1ex}
\caption{Analysis of FSP matrix distance as robustness indicator. We compute the FSP matrix distance for each cell, along with the performance gap between clean accuracy and adversarial accuracy. For cells in deeper positions of the network, the FSP distance has a positive correlation with the gap between network robustness and its clean accuracy, which indicates that a robust network has a lower FSP matrix loss in the deeper cells of the network.}
\vspace{1ex}
\label{appendix:fig-gram}
\end{figure*}

We provide additional results for the correlation of FSP matrix distance along with the performance gap between clean accuracy and adversarial accuracy in cell-free setting. Results for several cells have been shown in the main paper. Here we provide results for additional cells in Fig.~\ref{appendix:fig-gram}.

As can be observed from the figure, for cells in deeper positions of the network, the FSP distance has a positive correlation with the gap between network robustness and its clean accuracy, which indicates that a robust network has a lower FSP matrix loss in the deeper cells of the network.

\section{Visualization of RobNets}
In this section, we first describe the details of how we select architectures of RobNet family. Further, we visualize several representative RobNet architectures.

In cell-based setting, we first filter out the architectures with architecture density $D<0.5$. Then we only consider the architectures which have a portion of direct convolutions larger than $0.5$. For each of the computational budget, we sample 50 architectures from the supernet following the process described above and finetune them for 3 epochs to get the adversarial accuracy. We then select architectures with best performance under each computational budget, and refer to them as RobNet-small, RobNet-medium, and RobNet-large, respectively.

In cell-free setting, we first randomly sample 300 architectures from the supernet, and  calculate the average FSP matrix distance for last 10 cells of each sampled network.
Following the finding of FSP matrix loss as indicator, we reject those architectures whose average distance is larger than a threshold. In our experiments, we set the threshold to be $0.006$, which leads to 68 remaining architectures. Finally, we finetune each of them for 3 epochs and select the architecture with the highest adversarial accuracy, which is named as RobNet-free. 

\begin{figure*}
\centering
\includegraphics[width=.99\linewidth]{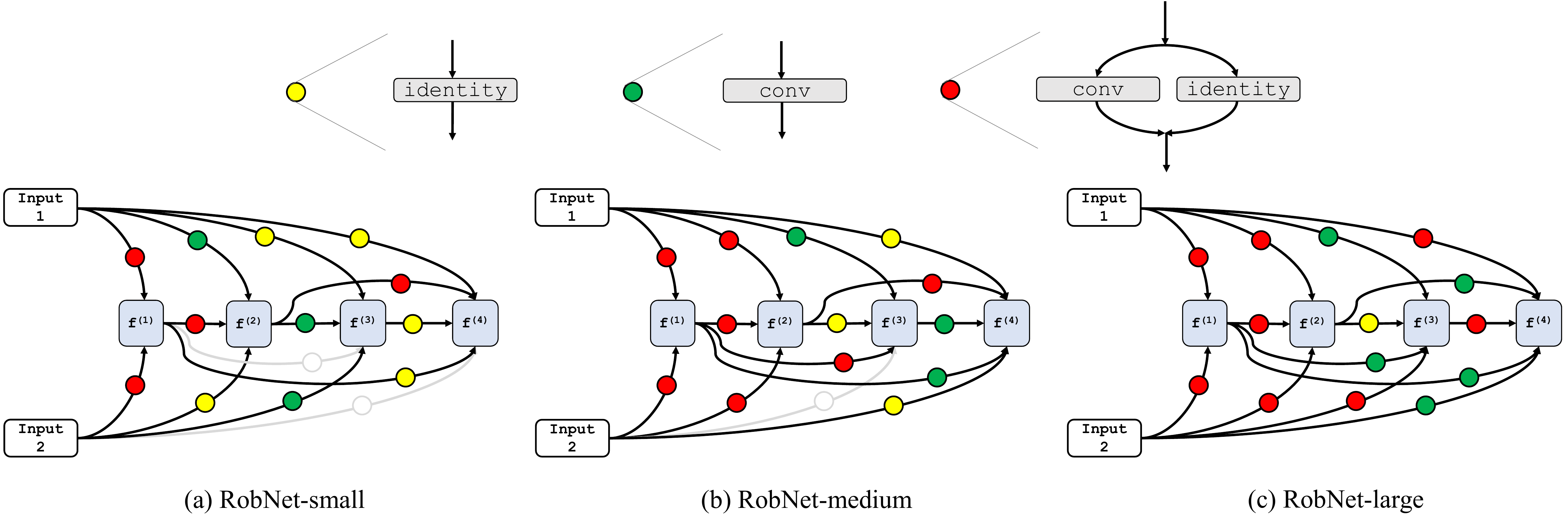}
\vspace{.5ex}
\caption{Visualization of representative architectures of RobNet family.}
% \vspace{-.3cm}
\label{appendix:fig-vis-arch}
\end{figure*}

We visualize several representative architectures of RobNet family in Fig.~\ref{appendix:fig-vis-arch}. 

% =========================================================================== %
\begin{table*}[!t]
\caption{Black-box PGD$^{100}$ attack results on CIFAR-10. All models are adversarially trained using PGD with $7$ steps. We create PGD adversarial examples with $\epsilon=8/255$~(0.031) for $100$ iterations from the evaluation set on the source network, and then evaluate them on an independently initialized target network.
The best results of each column are in \textbf{bold} and the empirical lower bound (the lowest accuracy of each row) for each network is \underline{underlined}.
}
\vspace{1ex}
\centering
\scalebox{0.95}{
\setlength{\tabcolsep}{1.5mm}{
\begin{tabular}{@{}c|c|c|c|c|c|c@{}}
\toprule
\diagbox{Target}{Source} & ResNet-18        & ResNet-50        & \begin{tabular}[c]{@{}c@{}}WideResNet-\\ 28-10\end{tabular} & \begin{tabular}[c]{@{}c@{}}DenseNet-\\ 121\end{tabular}     & RobNet-large     & RobNet-free            \\ \midrule
ResNet-18        & {\ul 54.28\%}    & 54.49\%          & 56.44\%          & 57.19\%          & 55.57\%          & 59.37\%                \\ \midrule
ResNet-50        & 56.24\%          & {\ul 55.89\%}    & 56.38\%          & 58.31\%          & 57.22\%          & 60.19\%                \\ \midrule
WideResNet-28-10 & 57.89\%          & 57.96\%          & {\ul 55.68\%}    & 58.41\%          & 59.08\%          & 60.74\%                \\ \midrule
DenseNet-121     & 61.42\%          & 61.96\%          & 60.28\%          & {\ul 59.34\%}    & 60.03\%          & 59.96\%                \\ \midrule
RobNet-large     & 59.63\%          & 59.82\%          & 59.72\%          & 60.03\%          & {\ul 59.58\%}    & 60.73\%                \\ \midrule
RobNet-free      & \textbf{66.64\%} & \textbf{66.09\%} & \textbf{65.05\%} & \textbf{64.40\%} & \textbf{63.35\%} & {\ul \textbf{63.17\%}} \\ \bottomrule
\end{tabular}
}}
% \vspace{-.3cm}
\label{table:appendix black-box}
\end{table*}
% =========================================================================== %

\section{Additional Black-box Attack Results}
\label{appendix:black-box}
We provide additional results on transfer-based black-box attacks on CIFAR-10, across different network architectures. The black-box adversarial examples are generated from an independently trained copy of the network, by using \emph{white-box} attack on the victim network. We apply PGD-based black-box attacks with 100 iterations across different architectures, and report the result in Table~\ref{table:appendix black-box}. All models are adversarially trained using PGD with $7$ steps.

In the table, we highlight the best result of each column in \textbf{bold}, which corresponds to the most robust model against black-box adversarial examples generated from one specific source network. We also \underline{underline} the empirical lower bound for each network, which corresponds to the lowest accuracy of each row.

As the table reveals, RobNet-free model achieves the highest robust accuracy under transfer-based attacks from different sources. Furthermore, the most powerful black-box adversarial examples for each network~(i.e., the underlined value) are from source network that uses the same architecture as the target network.
Finally, by comparing the transferability between two network architectures~(e.g., RobNet-free $\to$ ResNet-18 \& ResNet-18 $\to$ RobNet-free), we can observe the following phenomena.
First, our RobNet models are more robust against black-box attacks transferred from other models. Moreover, our RobNet models can generate stronger adversarial examples for black-box attacks compared with other widely used models.

\section{Additional White-box Attack Results}
\label{appendix:white-box}
\begin{figure}[h]
\centering
\includegraphics[width=.91\linewidth]{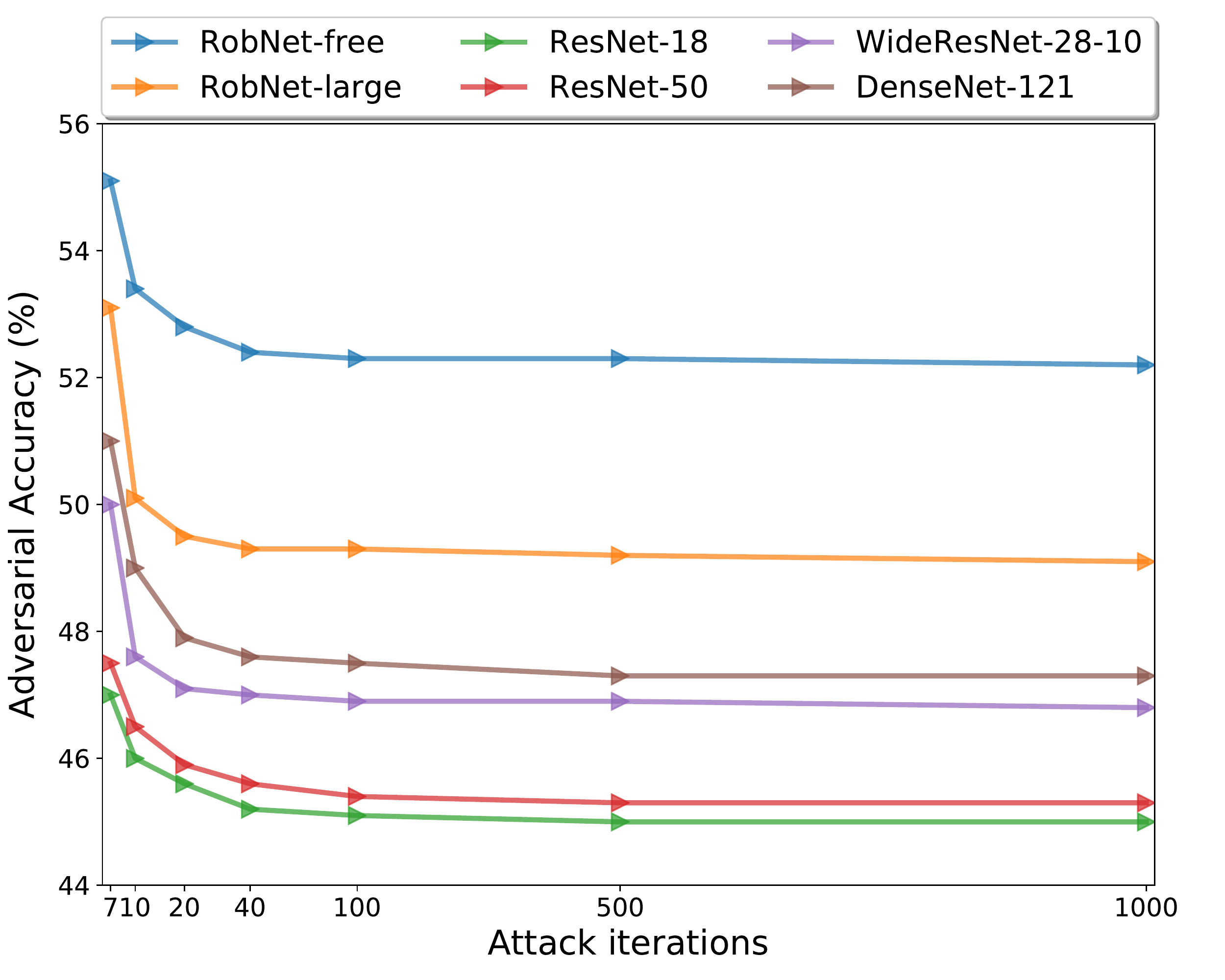}
% \vspace{-.2cm}
\caption{White-box attack results on CIFAR-10. All models are adversarially trained using PGD with $7$ steps. We show results of different architectures against a white-box PGD attacker with 7 to \textbf{1000} attack iterations.}
\label{appendix:fig-whitebox}
\end{figure}

% \noindent {\bf Attack Iterations.}
As common in recent literature~\cite{xie2019feature,yang2019menet,zhang2019theoretically}, \emph{strongest possible} attack should be considered when evaluating the adversarial robustness. Therefore, we further strengthen the adversary and vary the attack iterations from 7 to 1000.
We show the results in Fig.~\ref{appendix:fig-whitebox}, where RobNet family outperforms other networks, even facing the strong adversary.
Specifically, compared to state-of-the-art models, RobNet-large and RobNet-free can gain $\sim 2\%$ and $\sim 5\%$ improvement, respectively. We also observe that the attacker performance diminishes with 500$\sim$1000 attack iterations.

\vspace{.15cm}

\end{appendices}

\end{document}